\pdfoutput=1

\documentclass[11pt]{article}

\usepackage[final]{acl}

\usepackage{times}
\usepackage{latexsym}

\usepackage[T1]{fontenc}

\usepackage[utf8]{inputenc}

\usepackage{microtype}

\usepackage{inconsolata}
\usepackage{booktabs}

\usepackage{graphicx}
\usepackage{tcolorbox}
\usepackage{amsfonts}
\usepackage{amsmath}
\usepackage{xcolor}
\usepackage{multicol}
\usepackage{multirow}
\usepackage{subcaption}
\usepackage{colortbl}
\usepackage{enumitem}
\usepackage{wasysym}
\usepackage{makecell}
\usepackage{cleveref}
\usepackage{titletoc}
\usepackage{wasysym}
\usepackage{pifont}
\usepackage[toc,page,header]{appendix}
\usepackage{minitoc}
\usepackage{tikz}
\usetikzlibrary{mindmap,shapes, positioning}
\usepackage{smartdiagram}
\usesmartdiagramlibrary{additions}
\usepackage{forest}

\definecolor{blue-color}{RGB}{102, 153, 204}
\definecolor{harvestgold}{rgb}{0.85, 0.57, 0.0}

\title{Can LLM Watermarks Robustly Prevent Unauthorized Knowledge Distillation?}

\author{\textbf{Leyi Pan}$^{1}$, \textbf{Aiwei Liu}$^{1\dagger}$, \textbf{Shiyu Huang}$^{2}$, \textbf{Yijian Lu}$^{1}$, \textbf{Xuming Hu}$^{3}$, \\
\textbf{Lijie Wen}$^{1\dagger}$, \textbf{Irwin King}$^{4}$, \textbf{Philip S. Yu}$^{5}$\\
  $^1$Tsinghua University, $^2$Zhipu AI, \\ $^3$The Hong Kong University of Science and Technology (Guangzhou), \\
  $^4$The Chinese University of Hong Kong, $^5$University of Illinois at Chicago\\
\texttt{\small panly24@mails.tsinghua.edu.cn,}\\ \texttt{ \small liuaw20@mails.tsinghua.edu.cn, wenlj@tsinghua.edu.cn}
  }

\begin{document}
\maketitle
{
\let\thefootnote\relax\footnotetext{
$^\dagger$ Corresponding authors. }
}
\doparttoc
\faketableofcontents 

\maketitle
\begin{abstract}
The radioactive nature of Large Language Model (LLM) watermarking enables the detection of watermarks inherited by student models when trained on the outputs of watermarked teacher models, making it a promising tool for preventing unauthorized knowledge distillation. However, the robustness of watermark radioactivity against adversarial actors remains largely unexplored. In this paper, we investigate whether student models can acquire the capabilities of teacher models through knowledge distillation while avoiding watermark inheritance. We propose two categories of watermark removal approaches: pre-distillation removal through untargeted and targeted training data paraphrasing (UP and TP), and post-distillation removal through inference-time watermark neutralization (WN). Extensive experiments across multiple model pairs, watermarking schemes and hyper-parameter settings demonstrate that both TP and WN thoroughly eliminate inherited watermarks, with WN achieving this while maintaining knowledge transfer efficiency and low computational overhead. Given the ongoing deployment of watermarking techniques in production LLMs, these findings emphasize the urgent need for more robust defense strategies.
\end{abstract}

\section{Introduction}
\label{sec:intro}
\begin{figure}[h!]
    \centering
    \includegraphics[width=\linewidth]{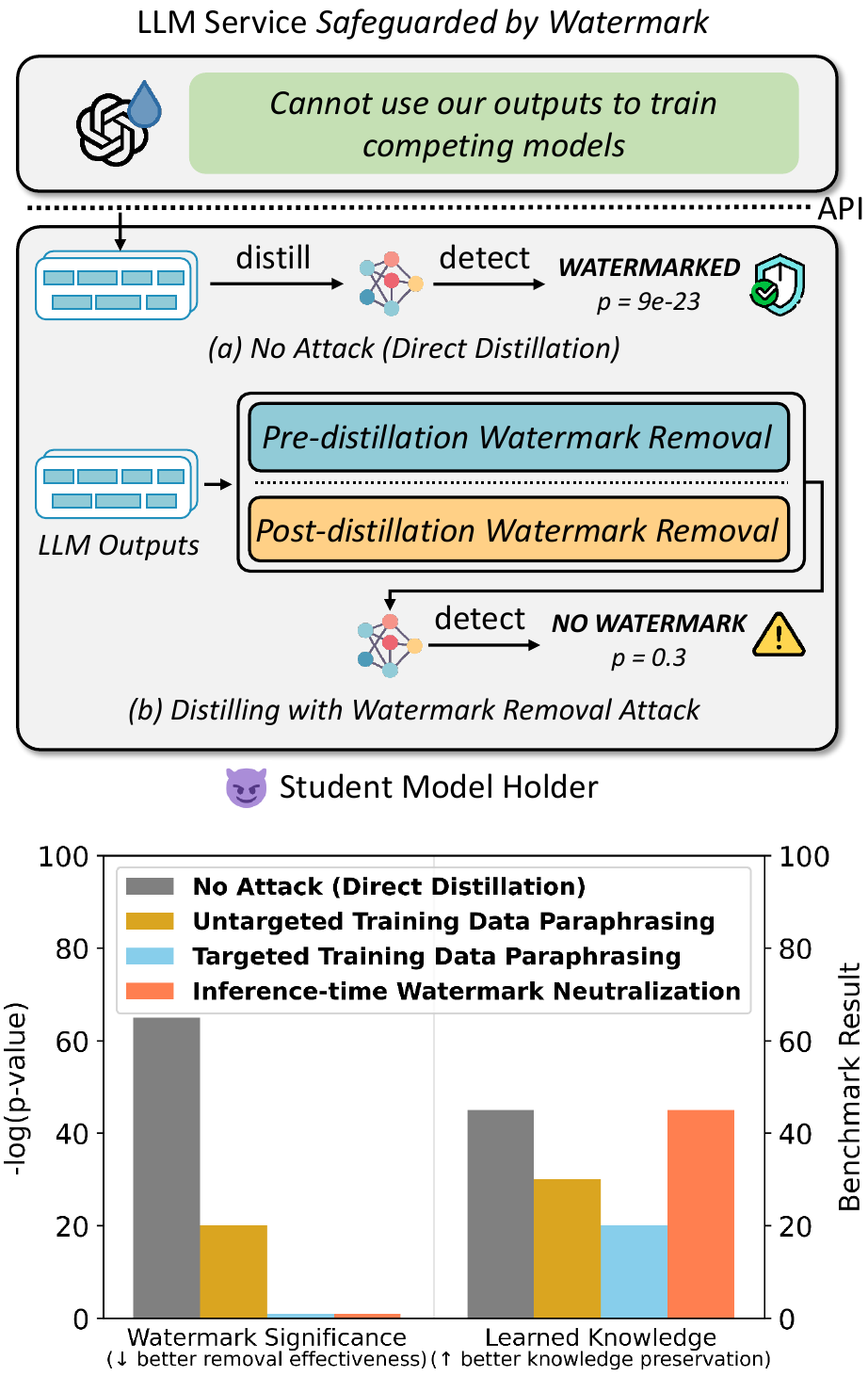}
    \caption{LLM watermarking has been proposed as a safeguard against unauthorized knowledge distillation. However, our pre- and post-distillation watermark removal attacks allow student models to perform untraceable knowledge distillation, emphasizing the need for more robust design. The bar chart displays the effectiveness of watermark removal and knowledge preservation for our three proposed attacks.}
    \label{fig:intro}
    \vspace{-20pt}
\end{figure}

The capability of Large Language Models (LLMs) to rapidly generate high-quality text at scale makes them valuable sources of training data \cite{zoph2022emergent}. However, many leading LLM services explicitly prohibit the use of their outputs for training competing models through knowledge distillation in their terms of service. Notable examples include OpenAI, Anthropic and Meta Llama, as detailed in Appendix \ref{sec:terms_of_use}. 

Watermarking has emerged as a solution to monitor unauthorized usage \cite{DBLP:conf/icml/KirchenbauerGWK23, zhao2023provable, liu2024a, zhao2023protecting}. Research has shown that watermarked LLMs exhibit radioactivity - student models trained on their outputs inherit detectable watermarks \cite{sander2024watermarking,gu2024learnability}. This traceability has led to increasing practical implementations, such as Google DeepMind's integration of SynthID-Text \cite{Dathathri2024} into Gemini chatbots \cite{team2023gemini}.

As watermarking emerges as a promising approach to protect model copyrights from knowledge distillation, its robustness against adversarial actors remains largely unexplored. We conduct the first systematic investigation into watermark resilience and propose two categories of watermark removal attacks: pre-distillation removal through untargeted and targeted training data paraphrasing (\textbf{UP} and \textbf{TP}), and post-distillation removal through inference-time watermark neutralization (\textbf{WN}), as illustrated in the upper part of Figure \ref{fig:intro}. Experiments show that TP and WN can thoroughly eliminate inherited watermarks, with WN \textbf{achieves watermark removal} while \textbf{preserving distilled knowledge} and \textbf{maintaining low computational overhead} - raising important questions about the reliability of preventing unauthorized knowledge distillation through watermarks.

Given that both TP and WN require knowledge of watermark rules, we propose a watermark stealing technique. Unlike existing methods \cite{jovanovicwatermark,wu-chandrasekaran-2024-bypassing,zhang2024large}, our approach (1) does not need access to the watermarking scheme or its hyper-parameters, and (2) assigns weights by analyzing factors affecting watermark radioactivity, allowing for more targeted rule extraction. In TP, we integrate the inverse of extracted watermark rules into paraphrase models like Dipper \cite{krishna2023paraphrasing} to remove watermark. In contrast, UP simply employs standard paraphrasing tools without considering rules. For post-distillation removal, we develop watermark neutralization that directly counteracts inherited watermarks by applying inverse rules during the student model's decoding phase.

Extensive experiments were conducted across 2 Teacher-Student model pairs $\times$ 2 leading watermarking schemes $\times$ 3 hyperparameter settings. The comparative results are summarized in the bottom part of Figure \ref{fig:intro}. Both TP and WN effectively eliminate inherited watermarks, reducing detection significance to levels similar to non-watermarked conditions (above $10^{-2}$) across all settings. Evaluations on benchmark datasets, including ARC challenge \cite{Clark2018ThinkYH}, TruthfulQA \cite{lin-etal-2022-truthfulqa}, and MTBench \cite{zheng2023judging} show that WN exhibits superior knowledge preservation, achieving comparable performance to baseline student models trained without any watermark removal techniques. This indicates that student models can leverage WN to remove watermarks without sacrificing model performance, posing a significant challenge to the practical deployment of watermark as a copyright protection mechanism.

\vspace{3pt}

\noindent \textbf{Key Contributions} \quad Our main contributions are:
\begin{itemize}
    \item We conduct the first systematic investigation into the robustness of watermarking schemes against adversarial actors in monitoring unauthorized knowledge distillation, proposing pre-distillation and post-distillation attacks.
    \item Our proposed targeted paraphrasing and watermark neutralization methods achieve thorough watermark removal, with the latter demonstrating superior knowledge preservation. This raises concerns about the reliability of current watermarking schemes for monitoring unauthorized knowledge distillation.
    \item Further discovery of watermark collisions in multi-source knowledge distillation scenarios reveals additional limitations of watermarking schemes in monitoring unauthorized knowledge distillation (Section \ref{sec:multi-source}). Given the ongoing deployment of watermarking techniques in production LLMs, these findings highlight the urgent need for more robust defense strategies (Section \ref{sec:defense}).
\end{itemize}

\section{Background}
\label{sec:background}
\begin{figure*}[t]
    \centering
    \includegraphics[width=\linewidth]{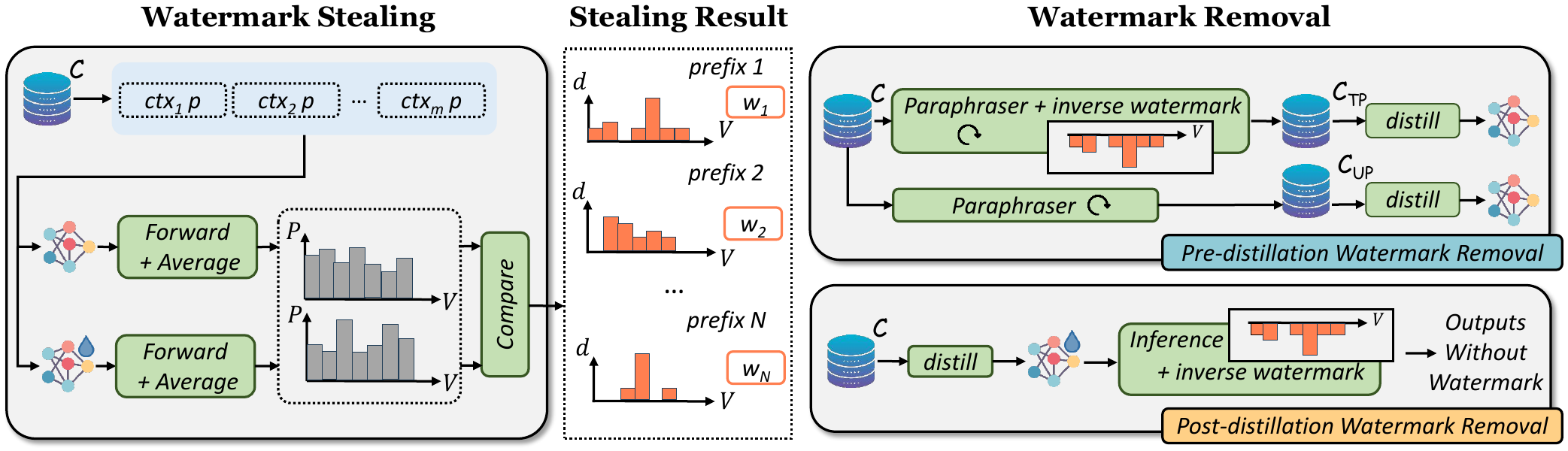}
    \caption{An illustration of the proposed watermark removal attacks. }
    \label{fig:method}
    \vspace{-10pt}
\end{figure*}
\subsection{LLM Watermarking Schemes} 
Most of the existing watermarking schemes follow the $n$-gram paradigm, modifying the next token's probability prediction based on the preceding $n-1$ tokens, thereby influencing the final sampling outcome \cite{DBLP:conf/icml/KirchenbauerGWK23, zhao2023provable,Dathathri2024,liu2024a,liu2024an,lee2023wrote,hu2023unbiased,wu2023dipmark,aronsonpowerpoint, kuditipudi2023robust, liu2024survey, lu2024entropy, pan-etal-2025-waterseeker}. 
Watermark schemes tested in this work are:

\textbf{KGW} \cite{DBLP:conf/icml/KirchenbauerGWK23} sets the ground work for generative LLM watermarking. For the $t^{th}$ token generation, it computes a hash $h_t = H(x_{t-n+1:t-1})$ from the previous $n-1$ tokens. This hash partitions the vocabulary $\mathcal{V}$ into a \textit{green list} $\mathcal{V}_g$ and a \textit{red list} $\mathcal{V}_r$. A constant bias $\delta$ is then added to the logits of green tokens:
\begin{equation}
\small
l'^{(i)}_t = l^{(i)}_t + \delta \text{\; if \;} v_i \in \mathcal{V}_g \text{\; else \;} l^{(i)}_t.
\end{equation}

As a result, watermarked text will statistically contain more \textit{green} tokens, and can be detected by computing the z-score:
\begin{equation}
\small
z = (|s|_G - \gamma T)/(\sqrt{\gamma(1 - \gamma) T}),
\end{equation}
where $|s|_G$ counts green tokens in text length $T$, and $\gamma = |\mathcal{V}_g|/|\mathcal{V}|$.

\vspace{3pt}

\textbf{SynthID-Text} \cite{Dathathri2024}, recently announced by Google DeepMind, is the first watermarking algorithm deployed in production, and has been integrated into the Gemini and Gemini Advanced chatbots. For the $t^{th}$ token generation, it computes a hash $h_t = H(x_{t-n+1:t-1})$ to seed $m$ binary classifiers $g_1, g_2, ..., g_m$, which randomly assign 0 or 1 to vocabulary tokens. It then samples $2^m$ tokens from the original distribution $P(x_t|x_{1:t-1})$ and conducts tournament sampling: tokens compete in pairs based on $g_1$ values in the first round, with subsequent rounds using $g_2, g_3, ..., g_m$ until one token remains. The watermark manifests as a statistical bias toward tokens with higher $g$ values, detectable by computing their mean: 
\begin{equation}
\small
    \overline{g} = \sum_{t=1}^T\sum_{\ell=1}^m g_\ell(x_t) / mT.
\end{equation}

\subsection{Watermark Radioactivity}
Research has shown that watermarked LLMs exhibit "radioactivity" — their distinctive watermark patterns can be inherited by student models trained on their outputs \cite{sander2024watermarking,gu2024learnability}. This characteristic is especially valuable for detecting unauthorized knowledge distillation. If a more capable teacher model employs watermark, any training data collected from its API will inherently carry the watermark. Consequently, student models trained on this data will produce outputs that retain the same watermark patterns, allowing the teacher model's owner to detect and trace unauthorized knowledge distillation attempts.

The underlying reason why ``the watermark has radioactivity" is that the watermark itself introduces a pattern that biases certain subsequent tokens based on the n-grams of the preceding context. When a student model is trained on watermarked data, it adopts this same bias. This effect is remarkably significant, with reported p-values below $10^{-30}$ even under the most stringent conditions: (1) the teacher model is closed-source, meaning student models can only learn from the output text and cannot access the teacher model's logits, which would otherwise make it easier to learn the pattern; and (2) watermark detection is performed unsupervised, meaning the test prompts for the student models are entirely disjoint from the training data.


\subsection{Watermark Removal Approaches}
Previous research has explored diverse approaches to watermark removal, primarily concentrating on eliminating watermarks from generated text rather than from the models themselves. These approaches can be categorized into untargeted and targeted methods. The untargeted approaches encompass various text transformation techniques, including paraphrasing, emoji-based attacks \cite{DBLP:conf/icml/KirchenbauerGWK23}, back-translation, and cross-lingual removal strategies \cite{he2024can}. In the realm of targeted removal, researchers such as \citet{jovanovicwatermark,wu-chandrasekaran-2024-bypassing} and \citet{zhang2024large} have proposed watermark stealing-and-removing techniques, though these methods necessitate prior knowledge of both the specific watermarking scheme employed and its corresponding window size parameter.


\section{Methodology}
\label{sec:method}
\subsection{Threat Model}

\noindent\textbf{Overview of the Attack Scenario} \quad Assume there is a closed-source teacher model, which employs a watermarking scheme, making all outputs acquired from its API carries the watermark. Now, consider the owner of a student model, who seeks to perform knowledge distillation by training on teacher's watermarked outputs. The owner applies various watermark removal attacks—such as untargeted training data paraphrasing, targeted training data paraphrasing, and watermark neutralization—with the goal of removing the inherited watermark from the trained student model, while preserving the knowledge it has learned.

\vspace{3pt}

\noindent\textbf{Watermarking Schemes Examined} \quad This investigation encompasses all n-gram-based watermarking schemes, including KGW \cite{DBLP:conf/icml/KirchenbauerGWK23}, SynthID-Text \cite{Dathathri2024}, KGW-Minhash \cite{kirchenbauer2023reliability}, KGW-SkipHash \cite{kirchenbauer2023reliability}, Unbiased Watermark\cite{hu2023unbiased}, DiPMark \cite{wu2023dipmark}, Aar \cite{aronsonpowerpoint}, and SIR \cite{liu2024a}, among others. These schemes represent the predominant paradigm of contemporary LLM watermarking methodologies. In our primary experiments, we have presented experimental findings for KGW and SynthID-Text. Furthermore, Appendix \ref{sec:more_schemes} contains supplementary results pertaining to KGW-Minhash and KGW-SkipHash.

\vspace{3pt}

\noindent\textbf{Access Requirements for the Student Model Owner to Perform Watermark Removal}\quad The student model owner requires the following: (1) Access to the training data (2) Access to both the original and trained student models. Since our removal methods are designed from the student model's perspective, these requirements are practical and feasible. Importantly, the student model owner does \textbf{NOT} need access to the watermark detection system or its API, which are controlled by the teacher model owner.

\noindent\textbf{Watermark Detection}\quad The watermark detection for the trained student model is performed by the owner of teacher model, involving prompting the trained student model to generate a certain amount of output text. These outputs are then analyzed to detect the presence of the watermark and to compute a confidence score. The test prompts are distinct from the training data, assuming that the teacher model service cannot track which specific data was used to train the student model.

\subsection{Overview of the Proposed Watermark Removal Methods}
We propose two categories of watermark removal methods: \textbf{pre-distillation} and \textbf{post-distillation} watermark removal, as illustrated in Figure \ref{fig:method}. Pre-distillation methods remove watermarks from training data using external paraphrase models. These methods include untargeted paraphrasing (\textbf{UP}), which directly rewrites training data, and targeted paraphrasing (\textbf{TP}), which first steals watermarking rules and then applies an inverse watermark on the paraphrase model to rewrite training data. Post-distillation method first steals watermark rules, and then neutralizes the inherited watermark by directly adding an inverse watermark during the student model's decoding phase. We refer to this process as watermark neutralization (\textbf{WN}). Details of these methods are presented in Sections \ref{sec:pre-distillation} and \ref{sec:post-distillation}, while our watermark stealing method used in both TP and WN is introduced in Section \ref{sec:watermark-stealing}.

\vspace{-3pt}

\subsection{Pre-distillation Watermark Removal} 
\label{sec:pre-distillation}
Let $\mathcal{R}$, $\mathcal{C}$, $\mathcal{O}$, and $\mathcal{W}$ denote the paraphrase model, training dataset collected from watermarked teacher model's API, original student model, and student model trained on $\mathcal{C}$ without attacks, respectively. For both TP and WN, we denote the watermark stealing result as $D(x_t; x_{t-n'+1:t-1})$, representing the confidence that $x_t$ is a watermarked token following $x_{t-n'+1:t-1}$. Section \ref{sec:watermark-stealing} details the computation of $D$.

\vspace{3pt}

\noindent\textbf{Targeted Training Data Paraphrasing} \quad During paraphrasing, we apply an inverse watermark to the paraphrase model $\mathcal{R}$'s logits based on $D$:
\begin{equation}
\small
{l'}_\mathcal{R}(x_t|x_{1:t-1}) = l_\mathcal{R}(x_t|{x_{1:t-1}}) - D(x_t; x_{t-n'+1:t-1}) \cdot \delta',
\end{equation}
where $\delta'$ controls the strength. This yields a new training dataset $\mathcal{C}_{TP}$ for the student model.

\vspace{3pt}

\noindent \textbf{Untargeted Training Data Paraphrasing} \quad As a comparison, this method directly applies $\mathcal{R}$ to rewrite training data, yielding dataset $\mathcal{C}_{UP}$.

\subsection{Post-distillation Watermark Removal}
\label{sec:post-distillation}
This approach neutralizes watermark by directly applying the inverse watermark to the trained student model $\mathcal{W}$'s logits during inference:
\begin{equation}
    \small
    {l'}_\mathcal{W}(x_t|x_{1:t-1}) = l_\mathcal{W}(x_t|{x_{1:t-1}}) - D(x_t; x_{t-n'+1:t-1}) \cdot \delta'.
\end{equation}

\vspace{-10pt}

\subsection{Watermark Stealing}
\label{sec:watermark-stealing}
This subsection presents our watermark stealing method that extracts token preferences following a prefix $p$ (denoted as $p$-rule).

\subsubsection{Watermark Radioactivity Factors}
\label{sec:factors}
\begin{figure}[t]
    \centering
    \includegraphics[width=0.85\linewidth]{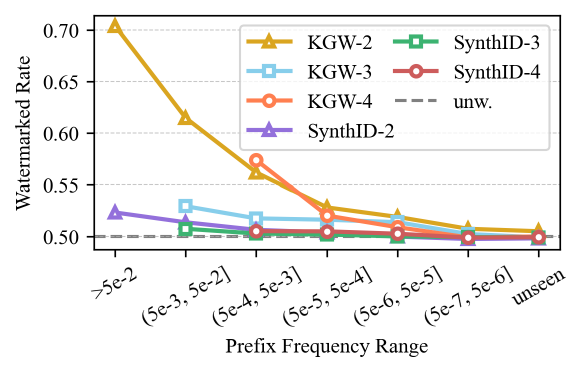}
    \caption{Correlation between prefix frequency in training data and the probability that tokens following these prefixes are watermarked in student model outputs.}
    \label{fig:freq_learnability}
\end{figure}
To efficiently steal watermarks, we first analyze factors that affecting watermark radioactivity. This analysis helps limit watermark stealing scope to rules with stronger inheritance patterns, reducing computational cost and minimizing model modifications needed for watermark removal. Our experiments reveal two key factors: \textbf{(1)} the occurring frequency of the prefix $p$ in training data, and \textbf{(2)} the window size $n$ used in watermarking schemes.

\vspace{3pt}

\noindent\textbf{Setup} \quad GLM-4-9b-chat \cite{glm2024chatglm} is used as the teacher model to generate 200k QA pairs for training Llama-7b \cite{touvron2023llama}. KGW \cite{DBLP:conf/icml/KirchenbauerGWK23} and SynthID-Text \cite{Dathathri2024} are used as watermarking schemes with $n=1,2,3,4$. We evaluated the inherited watermark strength in the student model using the C4 dataset \cite{raffel2020exploring} as prompts.

\vspace{3pt}


\noindent\textbf{Prefix Frequency vs. Radioactivity} \quad As shown in Figure \ref{fig:freq_learnability}, more frequent prefixes in training dataset lead to stronger watermark radioactivity of their $p$-rules in student model's outputs, across all schemes and settings. For rare prefixes (frequency $\le 5 \times 10^{-5}$), the radioactivity of their corresponding $p$-rules approaches that of unwatermarked text. Note: $n=1$ is excluded in the figure as it uses global, prefix-independent watermark rules.

\vspace{3pt}

\begin{figure}[t]
    \centering
    \vspace{10pt}
    \includegraphics[width=0.85\linewidth]{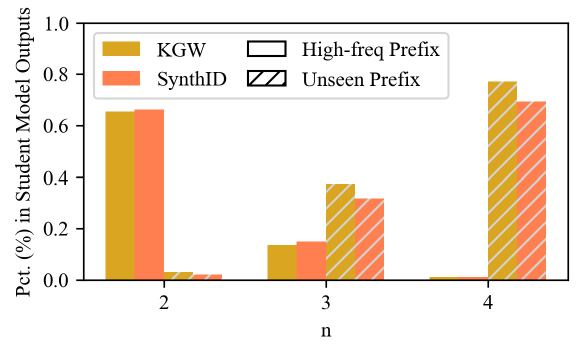}
    \caption{Percentages of high-frequency ($5\times 10^{-5}$) and unseen prefixes in training data within student model outputs, at different $n$.}
    \label{fig:prefix_num}
    \vspace{-10pt}
\end{figure}

\noindent\textbf{Window Size n vs. Radioactivity} \quad As shown in Table \ref{tab:window_size}, the watermark radioactivity falls dramatically as $n$ increases. For both KGW and SynthID-Text, watermarks become undetectable even with groups of \textit{1 million} tokens\footnote{For watermarked text, larger token samples yield stronger detection significance.} when $n$ reaches 4. This is because: (1) shorter $p$-rules are simpler, making it easier for student models to learn; (2) as $n$ increases, there is a marked expansion in the variety of prefixes generated by student models, resulting in fewer high-frequency prefixes and more unseen ones in the training data (as shown in Figure \ref{fig:prefix_num}). 

\begin{table}[t]
\caption{Median p-values for watermark detection in student model outputs, evaluated on groups of \textit{1 million} tokens, across varying watermark window sizes $n$.}
\centering
\resizebox{0.5\textwidth}{!}{
\begin{tabular}{ccccc}
    \toprule
    & $n=1$ & $n=2$ & $n=3$ & $n=4$ \\
    \midrule
    KGW & 6.24e-25979 & 4.79e-2537 & 1.67e-23 & 0.14 \\
    SynthID-Text & 6.20e-4028 & 6.08e-887 & 0.58 & 0.64 \\
    \bottomrule
\end{tabular}
}
\label{tab:window_size}
\vspace{-10pt}
\end{table}

\vspace{3pt}

\noindent\textbf{Scope of Watermark Stealing} \quad Based on the preceding analysis, when conducting watermark stealing, we need only focus on scenarios with small values of $n$ (i.e., $n \leq 3$). That is to say, \textit{prior knowledge of the window size employed in the teacher model's watermarking algorithm is not required}; instead, we can directly iterate through a small range of $n$ values, as detailed in Section \ref{sec:stealing}. Furthermore, for cases where $n \neq 1$, we can restrict our attention to high-frequency prefixes (i.e. those with frequencies exceeding $5 \times 10^{-5}$).

\subsubsection{Watermark Stealing Process}
\label{sec:stealing}
Unlike prior work \cite{jovanovicwatermark,wu-chandrasekaran-2024-bypassing,zhang2024large}, our proposed stealing method operates effectively without knowing the exact watermarking scheme or window size. We first assume a window size $n$ used by the teacher model to extract watermark rules, then obtain the final output by aggregating results from all windows less or equal to the maximum window size $n'$ considered. Based on Section \ref{sec:factors}, $n'$ typically remains small, ensuring manageable computational complexity.

\noindent\textbf{Scoring Single n-gram} \quad Regardless of the specific watermarking algorithm, the core mechanism is adjusting the sampling preferences of the subsequent token based on prefix tokens. Therefore, our objective is to identify preferred tokens following prefix $p=x_{t-n+1:t-1}$ by assigning a score in $[0,1]$ for each $v \in \mathcal{V}$, indicating the confidence value of ``$v$ is a watermarked token following $p$''.

Let $\mathcal{O}$ denote the original student model, $\mathcal{W}$ denote the student model after training on watermarked data, and $\mathcal{C}$ represent the training corpus. To extract $p$-rules, we collect all contexts in $\mathcal{C}$ that end with $p$, perform forward passes using both $\mathcal{O}$ and $\mathcal{W}$ on these contexts to obtain next token probability predictions, and average the predictions across different contexts, which are:
\begin{equation}
    \small
    \overline{P_\mathcal{O}}(x_t|p) = \mathbb{E}_{c \in \mathcal{C}, c_{t-n+1:t-1}=p}[P_\mathcal{O}(x_t | c)].
\end{equation}
\begin{equation}
    \small
    \overline{P_\mathcal{W}}(x_t|p) = \mathbb{E}_{c \in \mathcal{C}, c_{t-n+1:t-1}=p}[P_\mathcal{W}(x_t | c)].
\end{equation}
Comparing these two distributions reveals the context-independent statistical bias of tokens following prefix $p$, characterizing the watermark patterns. We quantify the distribution shift and score the $n$-gram using $d(x_t; x_{t-n+1:t-1})$:
\begin{equation}
    \small
    d(x_t; [x_{t-n+1:t-1}]) = \frac{1}{2} \min (2, \frac{\overline{P_W}(x_t|x_{t-n+1:t-1})}{\overline{P_\mathcal{O}}(x_t|x_{t-n+1:t-1})}),
\end{equation}
if $\overline{P_\mathcal{W}}(x_t|x_{t-n+1:t-1}) > \overline{P_\mathcal{O}}(x_t|x_{t-n+1:t-1})$. Otherwise, $\small d(x_t, x_{t-n+1:t-1})=0$. Note that if $n=1$, which means the watermark rule is globally fixed, $d(x_t)$ is computed by quantifying the average probability shifts across all contexts.

\vspace{3pt}

\noindent\textbf{Considering Multiple Window Sizes} \quad Since the window size $n$ of the watermark scheme used in the teacher model is unknown, we need to aggregate scoring results across different $n$-gram sizes. Let $n'$ be the maximum window size under consideration. The final confidence score is then defined as:
\begin{equation}
\small
\begin{aligned}
    D(x_t; x_{t-n'+1:t-1}) &= d(x_t) + \\
    &\sum_{i=1}^{n'-1} w(x_{t-i:t-1}) \cdot d(x_t; x_{t-i:t-1}),
\end{aligned}
\end{equation}
where $w(x_{t-i:t-1})$ is the weight assigned to the prefix based on its occurring frequency in training data. The weight value is computed as follows:
\begin{equation}
\small
w(x_{t-k:t-1}) = \begin{cases}
\left(\frac{\log f(x_{t-k:t-1})}{\log \max_{c \in \mathcal{C}_k} f(c)}\right)^{-\alpha} & \text{if } f(x_{t-k:t-1}) > \theta, \\
0 & \text{otherwise},
\end{cases}
\end{equation}
where $f$ denotes the occurring frequency in training data, $\mathcal{C}_k$ represents the set of all unique $k$-grams appearing in $\mathcal{C}$, and $\alpha$ is a smoothing parameter. This function assigns higher weight values to prefixes with higher frequency. 

\section{Experiments}
\begin{table*}[t]
    \caption{Median p-values for watermark detection using \textbf{UP} (Untargeted Training Data Paraphrasing), \textbf{TP} (Targeted Training Data Paraphrasing), and \textbf{WN} (Watermark Neutralization), compared against direct training (No Attack) and unwatermarked conditions (Unw.). \raisebox{0.5ex}{\colorbox{blue!30}{\quad}} indicates high watermark confidence, \raisebox{0.5ex}{\colorbox{blue!10}{\quad}} indicates low watermark confidence, and unshaded cells indicate insufficient evidence for watermark presence. Student model used in this table is Llama-7b, results for Llama-3.2-1b can be found in Appendix \ref{sec:remove_llama3_2}.}
    \centering
    \resizebox{0.90\textwidth}{!}{
    \begin{tabular}{cccccccc}
        \toprule
        \multicolumn{2}{c}{\textbf{Watermarking Scheme}} & \textbf{Token Num}. & \textbf{Unw.} & \textbf{No Attack} & \textbf{UP} & \textbf{TP} & \textbf{WN} \\
        \midrule
        \multirow{9}{*}{KGW}& \multirow{3}{*}{$n=1$} & 1k & 5.75e-01 & \cellcolor{blue!30}{8.97e-29} & \cellcolor{blue!10}6.82e-03 & 8.27e-01 & 8.20e-02 \\
        & & 2k & 5.71e-01 & \cellcolor{blue!30}6.49e-55 & \cellcolor{blue!10}2.43e-04 & 6.99e-01 & 2.72e-02 \\
        & & 3k & 6.01e-01 & \cellcolor{blue!30}2.68e-81 & \cellcolor{blue!30}9.68e-06 & 7.98e-01 & 1.21e-02 \\
        \cmidrule{2-8}
        & \multirow{3}{*}{$n=2$} & 10k & 4.80e-01 & \cellcolor{blue!30}4.12e-28 & \cellcolor{blue!10}2.18e-03 & 6.88e-01 & 9.85e-02 \\
        & & 20k & 4.47e-01 & \cellcolor{blue!30}4.12e-53 & \cellcolor{blue!10}1.29e-05 & 7.37e-01 & 3.35e-02 \\
        & & 30k & 3.62e-01 & \cellcolor{blue!30}8.26e-79 & \cellcolor{blue!30}1.05e-07 & 7.43e-01 & 1.24e-02 \\
        \cmidrule{2-8}
        & \multirow{3}{*}{$n=3$} & 100k & 3.40e-01 & \cellcolor{blue!10}1.85e-03 & 8.52e-01 & 4.30e-01 & 5.95e-01 \\
        & & 300k & 3.41e-01 & \cellcolor{blue!30}8.98e-09 & 9.51e-01 & 3.23e-01 & 6.80e-01 \\
        & & 1 million & 4.84e-01 & \cellcolor{blue!30}1.67e-23 & 8.63e-01 & 3.69e-01 & 8.63e-01 \\
        \midrule
        \multirow{9}{*}{SynthID-Text}& \multirow{3}{*}{$n=1$} & 1k & 9.67e-01 & \cellcolor{blue!10}1.46e-05 & 7.10e-01 & 9.98e-01 & 9.44e-01\\
        & & 2k & 9.95e-01 & \cellcolor{blue!30}1.08e-09 & 7.69e-01 & 9.96e-01 & 9.88e-01 \\
        & & 3k & 9.99e-01 & \cellcolor{blue!30}1.02e-13 & 8.05e-01 & 9.98e-01 & 9.97e-01 \\
        \cmidrule{2-8}
        & \multirow{3}{*}{$n=2$} & 10k & 4.23e-01 & \cellcolor{blue!30}6.67e-11 & 1.10e-01 & 4.97e-01 & 1.52e-01 \\
        & & 20k & 3.82e-01 & \cellcolor{blue!30}8.83e-20 & 4.29e-02 & 5.42e-01 & 7.30e-02 \\
        & & 30k & 3.09e-01 & \cellcolor{blue!30}1.65e-28 & 1.53e-02 & 4.45e-01 & 4.40e-02 \\
        \cmidrule{2-8}
        & \multirow{3}{*}{$n=3$} & 100k & 9.98e-01 & 5.28e-01 & 9.92e-01 & 9.94e-01 & 9.87e-01\\
        & & 300k & 9.76e-01 & 5.78e-01 & 9.99e-01 & 9.91e-01 & 9.49e-01\\
        & & 1 million & 9.87e-01 & 5.83e-01 & 9.99e-01 & 9.92e-01 & 9.85e-01\\
        \bottomrule
    \end{tabular}
    }
    \label{tab:removal}
\end{table*}
\subsection{Setup} 
\noindent\textbf{Teacher and Student Models} \quad Teacher: GLM-4-9b-chat \cite{glm2024chatglm}; Students: Llama-7b \cite{touvron2023llama} and Llama-3.2-1b \cite{dubey2024llama}. 

\vspace{3pt}

\noindent\textbf{Watermarking Schemes} \quad KGW \cite{DBLP:conf/icml/KirchenbauerGWK23} and SynthID-Text \cite{Dathathri2024} with $n=1,2,3$. Results for more watermarking schemes can be found in Appendix \ref{sec:more_schemes}. All watermarking algorithms tested are implemented using the MarkLLM toolkit \cite{pan2024markllm}. 

\vspace{3pt}

\noindent\textbf{Training Details} \quad Dataset is collected by prompting the teacher model to generate 200k QA pairs (detailed in Appendix \ref{sec:training-data}). We employ LlamaFactory \cite{zheng2024llamafactory} to perform supervised fine-tuning to the student models, with a learning rate of 1e-5 and 3 epochs for all test settings.

\vspace{3pt}

\noindent\textbf{Testing Details} \quad For watermark detection, we prompted the distilled student models to generate texts using C4 dataset \cite{raffel2020exploring}. The generated tokens were grouped into fixed-size samples, with p-values calculated for each group and the median reported. For knowledge preservation, we selected three representative benchmarks: ARC Challenge \cite{Clark2018ThinkYH} and TruthfulQA Multiple Choice \cite{lin-etal-2022-truthfulqa} (both multiple-choice tasks), along with the generative task MTBench \cite{zheng2023judging}. These benchmarks cover diverse areas including humanity, STEM, reasoning, writing, math, and coding.

\vspace{3pt}

\noindent\textbf{Others} \quad Frequency threshold $\theta=5\times 10^{-5}$, $n'=3$, smoothing parameter $\alpha=0.3$, inverse watermark strength $\delta'=2.5$ (adaptive control strategy for $\delta'$ can be found in Appendix \ref{sec:adaptive}). We use Dipper \cite{krishna2023paraphrasing} as the paraphraser.

\subsection{Effectiveness of Watermark Removal}

\textbf{Main Results} \quad Table \ref{tab:removal} demonstrates  the effectiveness of the three proposed watermark removal methods across different settings. It is evident that both TP and WN methods successfully eliminate the inherited watermark in all cases, maintaining confidence levels similar to unwatermarked conditions. The UP method also contributes to watermark removal; however, due to its lack of specificity, it fails to achieve complete removal when the watermark learned by the student model is strong (i.e., KGW $n=1,2$).

\vspace{3pt}

\begin{figure}[t]
    \centering
    \includegraphics[width=\linewidth]{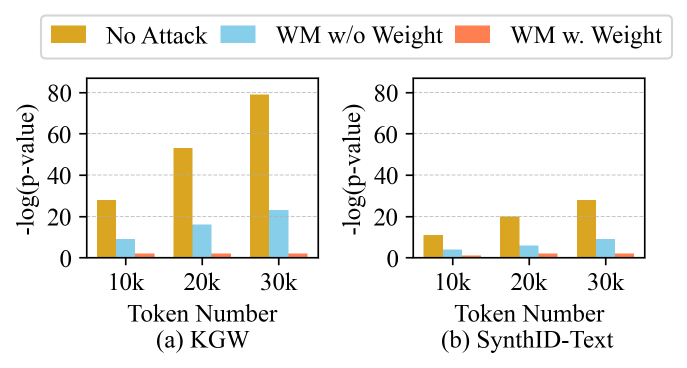}
    \caption{Comparison of watermark removal effectiveness: frequency-based prefix weighting vs. uniform weighting strategies.}
    \label{fig:weight}
\end{figure}

\noindent\textbf{Weight Ablation Study} \quad Figure \ref{fig:weight} compares watermark removal effectiveness of WN between frequency-based and uniform prefix weighting (using $n=2$). The results show that frequency-based prefix weighting, which assigns higher weights to more easily learned $p$-rules, achieves better watermark removal while maintaining an equal total weight across prefixes.

\begin{table*}[t]
    \caption{Comparison of student model performance across benchmarks under different scenarios: no attack (Trained SM), UP, TP and WN. Values in () indicate percentage changes relative to Trained SM, with the highest performance in each setting bolded and underlined. Student model used in this table is Llama-7b, results for Llama-3.2-1b can be found in Appendix \ref{sec:knowledge_llama3_2}.}
    \centering
    \resizebox{0.98\textwidth}{!}{
    \begin{tabular}{cccccccc}
        \toprule
        \textbf{Benchmark} &
        \textbf{Ori. SM} & \multicolumn{2}{c}{\textbf{Wat. Scheme}} & \textbf{Trained SM} & \textbf{Trained SM + UP} & \textbf{Trained SM + TP} &\textbf{Trained SM + WN} \\
        \midrule
        \multirow{6}{*}{\makecell{ARC \\ Challenge \\ (ACC)}} & \multirow{6}{*}{0.4181} & \multirow{3}{*}{KGW} & $n=1$ & 0.4480 & 0.4215 \textcolor[rgb]{0.5,0.0,0.0}{(-5.9\%)} & 0.3951 \textcolor[rgb]{0.5,0.0,0.0}{(-11.8\%)}& \underline{\textbf{0.4497}} \textcolor[rgb]{0.0,0.5,0.0}{(+0.6\%)}\\
        & & & $n=2$ & \underline{\textbf{0.4404}} & 0.4283 \textcolor[rgb]{0.5,0.0,0.0}{(-2.7\%)}& 0.4104 \textcolor[rgb]{0.5,0.0,0.0}{(-6.8\%)}& 0.4369 \textcolor[rgb]{0.5,0.0,0.0}{(-0.8\%)}\\
        & & & $n=3$ & \underline{\textbf{0.4778}} & 0.3865 \textcolor[rgb]{0.5,0.0,0.0}{(-19.1\%)}& 0.3840 \textcolor[rgb]{0.5,0.0,0.0}{(-19.6\%)}& 0.4642 \textcolor[rgb]{0.5,0.0,0.0}{(-2.8\%)}\\
        \cmidrule{3-8}
        & & \multirow{3}{*}{\makecell{SynthID\\ -Text}} & $n=1$ & 0.4505 & 0.4394 \textcolor[rgb]{0.5,0.0,0.0}{(-2.5\%)} & 0.4198 \textcolor[rgb]{0.5,0.0,0.0}{(-6.8\%)}& \underline{\textbf{0.4548}} \textcolor[rgb]{0.0,0.5,0.0}{(+1.0\%)} \\
        & & & $n=2$ & 0.4360 & 0.4403 \textcolor[rgb]{0.0,0.5,0.0}{(+1.0\%)}& 0.4241 \textcolor[rgb]{0.5,0.0,0.0}{(-2.7\%)}& \underline{\textbf{0.4565}} \textcolor[rgb]{0.0,0.5,0.0}{(+4.7\%)}\\
        & & & $n=3$ & \underline{\textbf{0.4505}} & 0.4394 \textcolor[rgb]{0.5,0.0,0.0}{(-2.5\%)}& 0.4283 \textcolor[rgb]{0.5,0.0,0.0}{(-4.9\%)}& 0.4471 \textcolor[rgb]{0.5,0.0,0.0}{(-0.8\%)}\\

        \midrule
        \multirow{6}{*}{\makecell{TruthfulQA \\ Multiple Choice \\ (ACC)}} & \multirow{6}{*}{0.3407} & \multirow{3}{*}{KGW} & $n=1$ & 0.3884 & 0.3917 \textcolor[rgb]{0.0,0.5,0.0}{(+0.8\%)}& 0.3785 \textcolor[rgb]{0.5,0.0,0.0}{(-2.5\%)} & \underline{\textbf{0.4186}} \textcolor[rgb]{0.0,0.5,0.0}{(+7.8\%)}\\
        & & & $n=2$ & \underline{\textbf{0.4376}} & 0.4097 \textcolor[rgb]{0.5,0.0,0.0}{(-6.4\%)} & 0.4089 \textcolor[rgb]{0.5,0.0,0.0}{(-6.6\%)}& 0.4353 \textcolor[rgb]{0.5,0.0,0.0}{(-0.5\%)}\\
        & & & $n=3$ & 0.4459 & 0.4315 \textcolor[rgb]{0.5,0.0,0.0}{(-3.2\%)}& 0.4055 \textcolor[rgb]{0.5,0.0,0.0}{(-9.1\%)}& \underline{\textbf{0.4632}} \textcolor[rgb]{0.0,0.5,0.0}{(+3.9\%)}\\
        \cmidrule{3-8}
        & & \multirow{3}{*}{\makecell{SynthID\\ -Text}} & $n=1$ & 0.4063 & 0.3780 \textcolor[rgb]{0.5,0.0,0.0}{(-7.0\%)} & 0.3597 \textcolor[rgb]{0.5,0.0,0.0}{(-11.5\%)} & \underline{\textbf{0.4262}} \textcolor[rgb]{0.0,0.5,0.0}{(+4.9\%)} \\
        & & & $n=2$ & 0.3991 & 0.3965 \textcolor[rgb]{0.5,0.0,0.0}{(-0.7\%)}& 0.4043 \textcolor[rgb]{0.0,0.5,0.0}{(+1.3\%)} & \underline{\textbf{0.4281}} \textcolor[rgb]{0.0,0.5,0.0}{(+7.3\%)}\\
        & & & $n=3$ & 0.4102 & 0.4009 \textcolor[rgb]{0.5,0.0,0.0}{(-2.3\%)}& 0.4062 \textcolor[rgb]{0.5,0.0,0.0}{(-1.0\%)}& \underline{\textbf{0.4330}} \textcolor[rgb]{0.0,0.5,0.0}{(+5.3\%)}\\
        
        \midrule
        \multirow{6}{*}{\makecell{MTBench \\ (Full Score: 10)}} & \multirow{6}{*}{2.64} & \multirow{3}{*}{KGW} & $n=1$ & \underline{\textbf{3.86}} & 3.04 \textcolor[rgb]{0.5,0.0,0.0}{(-21.2\%)}& 2.76 \textcolor[rgb]{0.5,0.0,0.0}{(-28.5\%)} & 3.67 \textcolor[rgb]{0.5,0.0,0.0}{(-4.9\%)}\\
        & & & $n=2$ & 3.99 & 3.40 \textcolor[rgb]{0.5,0.0,0.0}{(-14.8\%)}& 2.94 \textcolor[rgb]{0.5,0.0,0.0}{(-26.3\%)}& \underline{\textbf{4.02}} \textcolor[rgb]{0.0,0.5,0.0}{(+0.7\%)}\\
        & & & $n=3$ & \underline{\textbf{4.11}} & 3.27 \textcolor[rgb]{0.5,0.0,0.0}{(-20.4\%)}& 3.04 \textcolor[rgb]{0.5,0.0,0.0}{(-26.0\%)}& 3.99 \textcolor[rgb]{0.5,0.0,0.0}{(-2.9\%)}\\
        \cmidrule{3-8}
        & & \multirow{3}{*}{\makecell{SynthID\\ -Text}} & $n=1$ & \underline{\textbf{4.14}} & 3.27 \textcolor[rgb]{0.5,0.0,0.0}{(-21.0\%)}& 2.01 \textcolor[rgb]{0.5,0.0,0.0}{(-51.4\%)}& 4.13 \textcolor[rgb]{0.5,0.0,0.0}{(-0.2\%)} \\
        & & & $n=2$ & \underline{\textbf{4.24}} & 3.05 \textcolor[rgb]{0.5,0.0,0.0}{(-28.1\%)} & 2.84 \textcolor[rgb]{0.5,0.0,0.0}{(-33.0\%)} & 4.12 \textcolor[rgb]{0.5,0.0,0.0}{(-2.8\%)}\\
        & & & $n=3$ & \underline{\textbf{4.24}} & 2.90 \textcolor[rgb]{0.5,0.0,0.0}{(-31.6\%)}& 2.69 \textcolor[rgb]{0.5,0.0,0.0}{(-36.6\%)}& 4.16 \textcolor[rgb]{0.5,0.0,0.0}{(-1.9\%)}\\
        \bottomrule
    \end{tabular}
    }
    \label{tab:knowledge}
\end{table*}

\subsection{Performance of Knowledge Preservation}
\textbf{Main Results} \quad Table \ref{tab:knowledge} shows that training on 200k watermarked teacher samples significantly improves the student model's performance across all benchmarks (Trained SM vs Ori. SM), regardless of watermarking scheme or window size $n$. When applying removal methods, UP and TP generally degrade performance, especially on generative tasks like MTBench, with TP showing larger degradation than UP. WN effectively maintains knowledge - compared to the trained SM, it improves performance in about half the cases and shows minor decreases (under 5\%) in others, performing similarly to direct training without attacks.

\vspace{3pt}
\begin{figure}[t]
    \centering
    \includegraphics[width=\linewidth]{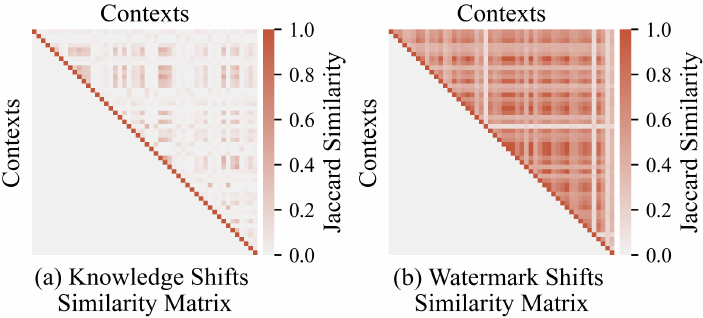}
    \caption{Pairwise similarities of probability prediction shifts across different contexts when the last token is fixed as ``the'', showing (a) knowledge shifts similarity and (b) watermark shifts similarity.}
    \label{fig:heatmap}
    \vspace{-10pt}
\end{figure}

\noindent\textbf{Why WN Can Achieve Good Knowledge Preservation} \quad WN achieves superior knowledge preservation for two key reasons. First, it eliminates the need for external rewriting tools. Since the teacher model represents our highest quality data source, avoiding external rewriting prevents data quality degradation. Second, our experiments reveal distinct patterns between knowledge and watermark learning. Knowledge learning depends on broader context, while watermark learning only relies on the previous $n-1$ tokens. Using $n=2$, we analyze 1,000 distinct text segments ending with "the" (\textit{ctx the}) and compare:
(1) \textbf{Knowledge shifts}: Probability differences between models before and after training on non-watermarked data show high variation across contexts (Figure \ref{fig:heatmap}(a)); (2) \textbf{Watermark shifts}: Probability differences between models trained on non-watermarked and watermarked data (generated using the same teacher and prompts) exhibit high consistency across different \textit{ctx} when the last token is fixed (Figure \ref{fig:heatmap}(b)).

Therefore, using prompts with fixed ending tokens to get averaged probability shifts mainly captures watermark patterns, while knowledge-related shifts tend to cancel out during averaging, resulting in minimal impact.

\section{Further Analysis}
\subsection{Computational Overhead}
\label{sec:computational}
To conduct a thorough assessment of practical applications, this section presents a detailed analysis of the operational overhead associated with the three proposed watermark removal methods, as summarized in Table \ref{tab:overhead_comparison}.

\vspace{3pt}

\noindent\textbf{External Tools} \quad Both UP (Untargeted Paraphrasing) and TP (Targeted Paraphrasing) require external paraphrasing tools for text transformation, which introduces additional dependencies and potential costs. In contrast, WN (Watermark Neutralization) operates independently without the need for external tools, making it more self-contained and easier to deploy.

\vspace{3pt}

\noindent\textbf{Preprocessing Time Consumption} \quad The preprocessing requirements vary significantly among the three methods. UP necessitates the use of Dipper-like models to paraphrase the entire training dataset, which can be computationally intensive and time-consuming. WN's preprocessing is more efficient, requiring only watermark stealing, which involves student model training (approximately 1 hour for Llama-7b) and dataset forward passes. TP combines both requirements, making it the most time-intensive in preprocessing. Notably, the forward passes required for watermark stealing are computationally more efficient than paraphrasing operations, as they can leverage parallel computation rather than relying on slower autoregressive generation.

\vspace{3pt}

\noindent\textbf{Student Model Inference Latency} \quad From an inference perspective, UP and TP maintain the original model's performance with no additional overhead during inference. WN introduces a minor latency due to the necessity of adding inverse watermark during inference. However, this additional computational cost is relatively minimal and may be acceptable in many practical applications.

\begin{table}[t]
\centering
\caption{Overhead comparison of watermark removal methods, averaged across various watermark settings using Llama-7b as the student model. All experiments were conducted on 8 NVIDIA H800 GPUs.}
\small
\resizebox{0.47\textwidth}{!}{
\begin{tabular}{lccc}
\toprule
& \textbf{UP} & \textbf{TP} & \textbf{WN} \\
\midrule
\textbf{Required External Tools} & \text{\ding{51}} & \text{\ding{51}} & \text{\ding{55}} \\
\textbf{Preprocessing Time (h)} & 31.8 & 37.1 & 3.6 \\
\textbf{Inference Latency (s / token)} & 0.0000 & 0.0000 & 0.0068 \\
\bottomrule
\end{tabular}
}
\label{tab:overhead_comparison}
\end{table}

\subsection{Multi-Source Knowledge Distillation}
\begin{table}[t]
    \caption{Watermark detection results in multi-source settings: 2 teacher models (1) employing KGW scheme with opposing keys; (2) using KGW and SynthID-Text respectively. Single-source results are shown in \textcolor{gray}{()}.}
    \centering
    \resizebox{0.47\textwidth}{!}{
    \begin{tabular}{ccc}
        \toprule
        & \textbf{KGW $k$ + KGW $\overline{k}$} & \textbf{KGW + SynthID-Text} \\
        \midrule
        Detector 1 & 2.81e-01 \textcolor{gray}{(4.21e-28)} & 5.64e-09 \textcolor{gray}{(4.21e-28)}  \\
        Detector 2 & 7.19e-01 \textcolor{gray}{(4.21e-28)} & 3.48e-03 \textcolor{gray}{(6.67e-11)}  \\
        \bottomrule
    \end{tabular}
    }
    \label{tab:multi-source}
    \vspace{-10pt}
\end{table}
\begin{figure}[t]
    \centering
    \includegraphics[width=\linewidth]{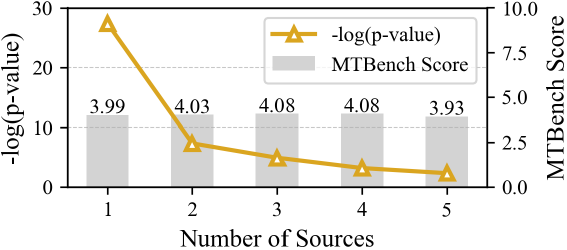}
    \caption{Average -$\log p$ values from detectors and MTBench scores as the number of teacher sources increases.}
    \label{fig:multi_source}
    \vspace{-15pt}
\end{figure}
\label{sec:multi-source}
The previous analysis focused on single-source knowledge distillation. In real-world scenarios with multiple LLM services, we found that watermarks from different sources can collide and counteract each other during knowledge distillation, making watermarks less effective as a protection mechanism - even without any removal methods. 

\vspace{3pt}

\noindent\textbf{Case 1: Two Opposing Keys} \quad We tested an extreme case with two teachers using the KGW scheme with complementary  keys (the hash results were complete opposites). When we trained a student model using a combined dataset (100k samples from each teacher), the watermark detection confidence dropped significantly from e-28 to e-01, as shown in Table \ref{tab:multi-source}.

\vspace{3pt}

\noindent\textbf{Case 2: Two Watermarking Schemes} \quad We examined a scenario using two teacher models with different watermarking methods (KGW and SynthID-Text), each generating 100k samples. Training a student model on this combined dataset led to reduced watermark detection confidence for both schemes' detectors compared to single-source scenarios (as shown in Table \ref{tab:multi-source}).

\vspace{3pt}

\noindent\textbf{Case 3: Multi-Source} \quad In this scenario, all teacher models employ the KGW scheme with randomly selected keys. As shown in Figure \ref{fig:multi_source}, while increasing the number of source models and maintaining a constant total volume of mixed training data, watermark detection became increasingly difficult, yet model performance remained stable. This suggests that mixing data from a sufficient number of teacher sources can achieve untraceable knowledge distillation.

\vspace{-3pt}

\subsection{Future Directions in Defense Strategy}
\label{sec:defense}
This work reveals the vulnerability of unauthorized knowledge distillation prevention when generative LLM watermarking is predominantly confined to n-gram based approaches. While alternatives such as sentence-level reject sampling \cite{hou2023semstamp,hou-etal-2024-k} and post-generation signal embedding \cite{chang2024postmark} exist, these approaches introduce significant latency to the current token-by-token real-time LLM inference paradigm, making them difficult to deploy at scale in real-time LLM services. 

Given these constraints, we advocate for a more diverse ecosystem of token-level watermarking techniques. The development of multiple paradigms would create a more complex landscape for attackers to navigate, while maintaining the efficiency required for real-world applications. Future research should focus on novel token-level approaches that can be seamlessly integrated into existing LLM inference pipelines while providing robust protection against various forms of attacks.
\vspace{-12pt}
\section{Conclusion}
This work presents the first systematic study of the robustness of watermarking schemes against adversarial attacks in preventing unauthorized knowledge distillation. We propose three watermark removal approaches: two pre-distillation methods (\textbf{U}ntargeted \textbf{P}araphrasing, \textbf{T}argeted \textbf{P}araphrasing) and one post-distillation method (\textbf{W}atermark \textbf{N}eutralization). Through comprehensive experiments, we evaluate the resilience of watermarking schemes against these attacks. Our findings reveal that WN achieves effective watermark removal while maintaining superior knowledge preservation, highlighting the urgent need for more robust defensive strategies.
\section*{Limitations}
While our study presents a systematic investigation of watermark resilience against adversarial attacks under the scenario of preventing unauthorized knowledge distillation, there still exist several limitations. Due to computational constraints, we only evaluated one teacher model (GLM-4-9b-chat) and two student models (Llama-7b and Llama-3.2-1b). The experiments were conducted using a fixed training dataset size of 200,000 samples and tested primarily on English language tasks. Additionally, our evaluation metrics focused on standard benchmarks (ARC, TruthfulQA, MTBench) and may not fully reflect performance on specialized domain tasks. Future work could explore a broader range of model architectures, training data scales, and task domains.
\section*{Acknowledgments}
This work is primarily supported by the Key Research and Development Program of China (No. 2024YFB3309702). We would like to express our gratitude to the anonymous ARR Feburary reviewers (Reviewer mdgr, LhPY, 7gDE) and Area Chair LGVF for their valuable feedback and suggestions that helped improve this paper.

\bibliography{custom}

\newpage
\onecolumn
\appendix

\begin{center}
{\huge \bfseries Appendices}
\end{center}

\vspace{3em}

\noindent{\Large \bfseries Table of Contents}

\vspace{0.5em}

\hrule height 0.5pt  

\startcontents[appendix]
\printcontents[appendix]{}{1}{}

\vspace{0.7em}

\hrule height 0.5pt  

\newpage

\section{Selected Terms of Use for LLM Services}
\label{sec:terms_of_use}
Figure \ref{fig:terms} shows excerpts from terms of use across various leading LLM services, including OpenAI\footnote{https://openai.com/policies/terms-of-use/}, Anthropic\footnote{https://www.anthropic.com/legal/consumer-terms} and Meta Llama\footnote{https://ai.meta.com/llama/license/}. These terms explicitly prohibit using model outputs for training or improving other models.
\begin{figure}[h!]
\begin{tcolorbox}[colback=gray!10, colframe=black, rounded corners]
\textbf{OpenAI.} What you cannot do.\\[0.5em]
\textit{Use output to develop models that compete with OpenAI.} \\[0.5em]
\textbf{Anthropic.} You may not access or use, or help another person to access or use, our Services in the following ways: \\[0.5em] \textit{To develop any products or services that compete with our Services, including to develop or train any artificial intelligence or machine learning algorithms or models.} \\[0.5em]
\textbf{Meta (Llama 2).} License Rights and Redistribution.\\[0.5em]
\textit{You will not use the Llama Materials or any output or results of the Llama Materials to improve any other large language model (excluding Llama 2 or derivative works thereof)}.
\end{tcolorbox}
\caption{Selected terms of use for various LLM services.}
\label{fig:terms}
\end{figure}
\section{Details of Watermarking Schemes}
\label{sec:schemes}
\subsection{KGW}
\noindent\textbf{Watermarking}\quad KGW \cite{DBLP:conf/icml/KirchenbauerGWK23} is a fundamental scheme of LLM watermarking. For generating the $t$-th token, the algorithm examines the previous $n-1$ tokens: $x_{t-n+1:t-1}$. These tokens are fed into a hash function $H$ to produce $h_t = H(x_{t-n+1:t-1})$. Based on $h_t$, the vocabulary $\mathcal{V}$ is deterministically split into a \textit{green list} $\mathcal{V}_g$ and a \textit{red list} $\mathcal{V}_r$. A constant bias $\delta$ is applied to logits of green tokens according to:

\begin{equation}
{l'}_t^{(i)} = \begin{cases}
{l}_t^{(i)} + \delta & \text{if } v_i \in \mathcal{V}_g \\
{l}_t^{(i)} & \text{if } v_i \in \mathcal{V}_r
\end{cases}
\end{equation}

\noindent\textbf{Detection} \quad Given a text sequence of length $T$, we count the number of green tokens $|s|_G$. Let $\gamma = |\mathcal{V}_g|/|\mathcal{V}|$ represent the expected proportion of green tokens in random text. The statistical significance of the green token count is measured by the z-score:

\begin{equation}
z = \frac{|s|_G - \gamma T}{\sqrt{\gamma(1-\gamma)T}}
\end{equation}

For a fixed $\delta$ ($\delta > 0$), longer sequences lead to stronger detection signal, as the z-score increases with text length $T$. 

\begin{figure}[t]
\centering
\begin{subfigure}[t]{0.48\textwidth}
\centering
\vbox to 6cm{
\vfill
\includegraphics[width=\textwidth,height=6cm,keepaspectratio]{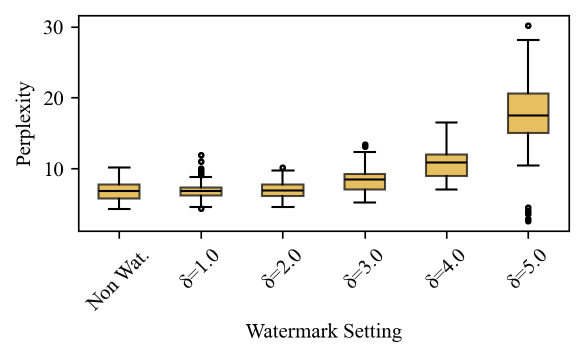}
\vfill
}
\caption{KGW}
\label{fig:perplexity_kgw}
\end{subfigure}
\hfill
\begin{subfigure}[t]{0.48\textwidth}
\centering
\vbox to 6cm{
\vfill
\includegraphics[width=\textwidth,height=6cm,keepaspectratio]{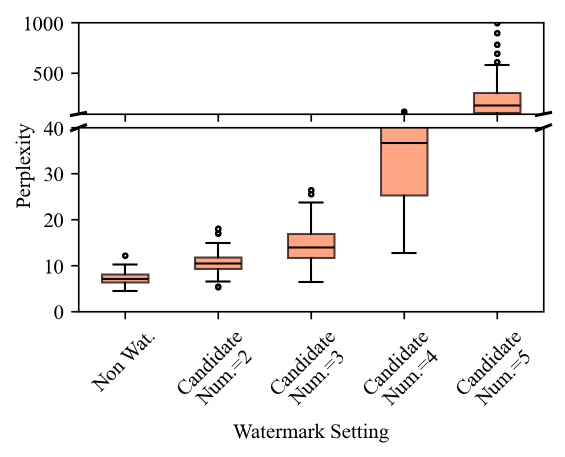}
\vfill
}
\caption{SynthID-Text}
\label{fig:perplexity_synthid}
\end{subfigure}
\caption{The relationship between watermark strength settings and perplexity under different watermarking schemes. The model used for calculating PPL is Llama-3.1-70B \cite{dubey2024llama}.}
\label{fig:perplexity}
\end{figure}

In our experiments, we set $\delta=3.0$ and $\gamma=0.5$, which represents a relatively strong watermark configuration in typical KGW settings. We avoid using larger $\delta$ values since stronger watermarks would notably degrade the text quality (as shown in Figure \ref{fig:perplexity_kgw}), making them impractical for real-world LLM services.

\noindent\textbf{Watermark Confidence: p-value} \quad Under the null hypothesis (non-watermarked text), the z-score follows a standard normal distribution $\mathcal{N}(0,1)$. The p-value can be computed as:
\begin{equation}
    p = 1-\Phi(z).
\end{equation}
Smaller p-value suggests higher confidence of watermark presence, as it indicates that the proportion of green tokens significantly exceeds what would be expected by chance.

\subsection{SynthID-Text}
\noindent\textbf{Watermarking} \quad SynthID-Text \cite{Dathathri2024} employs a tournament-based watermarking approach during the token generation process. For generating the $t$-th token, the algorithm first generates a random seed $h_t$ by applying a hash function $H$ to the previous $n-1$ tokens: $h_t = H(x_{t-n+1:t-1})$. This seed initializes $m$ independent $g$ functions $g_1, g_2, ..., g_m$, which assign binary values (0 or 1) to each token in the vocabulary.

The core watermarking process involves a tournament with $m$ layers. Initially, $2^m$ candidate tokens are sampled from the language model's original probability distribution $P(x_t|x_{1:t-1})$. These tokens undergo a series of pairwise competitions through $m$ tournament layers. In each layer $\ell$, tokens are randomly paired, and within each pair, the token with the higher $g_\ell$ score advances to the next layer, with random tie-breaking. The final surviving token after $m$ layers becomes the output token $x_t$.

\vspace{3pt}

\noindent\textbf{Detection} \quad Detection in SynthID-Text relies on measuring the statistical signature introduced during the watermarking process. Given a piece of text $x = x_1,...,x_T$, the detection algorithm:

\begin{enumerate}
\item Reconstructs the random seeds $h_t$ for each position $t$ using the same hash function and watermarking key
\item Computes the $g$-values for each token using the same watermarking functions  
\item Calculates the mean score across all positions and layers:
\end{enumerate}

\begin{equation}
\overline{g} = \frac{1}{mT} \sum_{t=1}^T \sum_{\ell=1}^m g_\ell(x_t).
\end{equation}

Due to the tournament selection process, watermarked text tends to contain tokens with higher $g$-values compared to non-watermarked text. Several factors contribute to a stronger detection signal: increasing the number of $g$ functions (larger $m$), using more candidates in each round of tournament sampling, and extending the sequence length for detection.

In our experiment, we set $m=30$ and use 2 candidates per round in tournament sampling, following the default configuration in SynthID-Text paper. We avoid using larger candidates number since stronger watermarks would notably degrade the text quality (as shown in Figure \ref{fig:perplexity_synthid}), making them impractical for real-world LLM services. However, we built upon this foundation by implementing the distortionary version of SynthID-Text, which enhances watermark strength. The non-distortionary version employs repeated context masking, where watermarks are only applied to subsequent tokens upon the first encounter with a particular prefix, while original sampling is used for subsequent occurrences of the same prefix. This approach ensures unbiased multi-step sampling and maintains text quality at the cost of watermark strength. In contrast, the distortionary version foregoes repeated context masking, sacrificing some text quality but achieving stronger watermarks.

\vspace{3pt}

\noindent\textbf{Watermark Confidence: p-value} \quad For a given text $x$ and its score $\overline{g}$, the $p$-value is calculated based on the following principles: Under the null hypothesis (text contains no watermark), $\overline{g}$ approximately follows a normal distribution according to the Central Limit Theorem. This distribution has a mean $\mu = 0.5$ since $g$ functions output 0 and 1 with equal probability for non-watermarked text. The variance $\sigma^2$ is estimated as $\frac{1}{4mT}$, where $m$ is the number of $g$ functions and $T$ is the text length. The $p$-value is then computed as:

\begin{equation}
p = 1 - \Phi(\frac{\overline{g} - 0.5}{1/\sqrt{4mT}})
\end{equation}

where $\Phi(\cdot)$ is the cumulative distribution function of the standard normal distribution. A smaller p-value indicates stronger evidence that the text contains a watermark.
\section{Experiment Results for Llama-3.2-1b}
This section provides supplementary experimental results for Llama-3.2-1b, including the effectiveness of watermark removal and knowledge preservation performance.
\subsection{Results of Watermark Removal Effectiveness for Llama-3.2-1b}
\label{sec:remove_llama3_2}
\begin{table*}[t]
    \caption{Median p-values for watermark detection using \textbf{UP} (Untargeted Training Data Paraphrasing), \textbf{TP} (Targeted Training Data Paraphrasing), and \textbf{WN} (Watermark Neutralization), compared against direct training (No Attack) and unwatermarked conditions (Unw.). \raisebox{0.5ex}{\colorbox{blue!30}{\quad}} indicates high watermark confidence, \raisebox{0.5ex}{\colorbox{blue!10}{\quad}} indicates low watermark confidence, and unshaded cells indicate insufficient evidence for watermark presence. Student model used in this table is Llama-3.2-1b.}
    \centering
    \resizebox{0.87\textwidth}{!}{
    \begin{tabular}{cccccccc}
        \toprule
        \multicolumn{2}{c}{\textbf{Watermarking Scheme}} & \textbf{Token Num}. & \textbf{Unw.} & \textbf{No Attack} & \textbf{UP} & \textbf{TP} & \textbf{WN} \\
        \midrule
        \multirow{9}{*}{KGW}& \multirow{3}{*}{$n=1$} & 1k & 5.75e-01 & \cellcolor{blue!30}1.26e-36 & \cellcolor{blue!10}3.27e-03 & 9.98e-01 & 6.21e-01 \\
        & & 2k & 5.71e-01 & \cellcolor{blue!30}3.59e-71 & \cellcolor{blue!10}6.00e-05 & 9.95e-01 & 7.00e-01 \\
        & & 3k & 6.01e-01 & \cellcolor{blue!30}1.28e-104 & \cellcolor{blue!30}1.03e-06 & 9.98e-01  & 7.34e-01 \\
        \cmidrule{2-8}
        & \multirow{3}{*}{$n=2$} & 10k & 4.80e-01 & \cellcolor{blue!30}4.79e-65 & \cellcolor{blue!10}1.00e-03 & 8.14e-01 & 5.20e-01 \\
        & & 20k & 4.47e-01 & \cellcolor{blue!30}5.96e-128 & \cellcolor{blue!30}6.42e-06 & 6.79e-01 & 4.97e-01 \\
        & & 30k & 3.62e-01 & \cellcolor{blue!30}4.68e-190 & \cellcolor{blue!30}3.59e-08 & 8.92e-01  & 5.48e-01 \\
        \cmidrule{2-8}
        & \multirow{3}{*}{$n=3$} & 100k & 3.40e-01 & \cellcolor{blue!30}6.85e-12 & 6.96e-02 & 1.98e-01 & 8.13e-01  \\
        & & 300k & 3.41e-01 & \cellcolor{blue!30}2.33e-27 & 1.09e-02 & 5.19e-01 & 9.43e-01 \\
        & & 1 million & 4.84e-01 & \cellcolor{blue!30}1.13e-87 & \cellcolor{blue!10}9.80e-05 & 2.04e-01 & 9.98e-01 \\
        \midrule
        \multirow{9}{*}{SynthID-Text}& \multirow{3}{*}{$n=1$} & 1k & 9.67e-01 & \cellcolor{blue!30}3.76e-10 & 2.55e-01 & 9.98e-01 & 9.94e-01 \\
        & & 2k & 9.95e-01 & \cellcolor{blue!30}9.00e-19 & 1.66e-01 & 9.87e-01 & 9.96e-01\\
        & & 3k & 9.99e-01 & \cellcolor{blue!30}1.72e-27 & 1.26e-01 & 9.89e-01 & 9.98e-01 \\
        \cmidrule{2-8}
        & \multirow{3}{*}{$n=2$} & 10k & 4.23e-01 & \cellcolor{blue!30}1.29e-32 & \cellcolor{blue!10}4.81e-03 & 2.93e-01 & 4.58e-01 \\
        & & 20k & 3.82e-01 & \cellcolor{blue!30}1.48e-63 & \cellcolor{blue!10}1.11e-04 & 1.83e-01 & 4.77e-02 \\
        & & 30k & 3.09e-01 & \cellcolor{blue!30}2.22e-95 & \cellcolor{blue!30}4.28e-06 & 3.67e-01 & 1.51e-02 \\
        \cmidrule{2-8}
        & \multirow{3}{*}{$n=3$} & 100k & 9.98e-01 & \cellcolor{blue!10}2.41e-05 & 9.93e-01 & 9.99e-01 & 8.95e-01 \\
        & & 300k & 9.76e-01 & \cellcolor{blue!30}2.47e-14 & 9.94e-01 & 9.99e-01 & 9.97e-01 \\
        & & 1 million & 9.87e-01 & \cellcolor{blue!30}1.85e-32 & 9.97e-01 & 9.99e-01 & 9.98e-01\\
        \bottomrule
    \end{tabular}
    }
    \label{tab:removal_llama3_2}
\end{table*}
Table \ref{tab:removal_llama3_2} demonstrates the effectiveness of the three proposed watermark removal approaches across various settings, with Llama-3.2-1b serving as the student model. The experimental results align closely with those obtained using Llama-7b as the student model, as discussed in the main text: both targeted training data paraphrasing (TP) and inference-time watermark neutralization (WN) successfully eliminate the watermark completely, while untargeted training data paraphrasing (UP) shows some removal effect but fails to achieve complete elimination across all scenarios.

\subsection{Results of Knowledge Preservation Performance for Llama-3.2-1b}
\label{sec:knowledge_llama3_2}
\begin{table*}[t]
    \caption{Comparison of student model performance across benchmarks under different scenarios: no attack (Trained SM), UP, TP and WN. Values in () indicate percentage changes relative to Trained SM, with the highest performance in each setting bolded and underlined. Student model used in this table is Llama-3.2-1b.}
    \centering
    \resizebox{0.98\textwidth}{!}{
    \begin{tabular}{cccccccc}
        \toprule
        \textbf{Benchmark} &
        \textbf{Ori. SM} & \multicolumn{2}{c}{\textbf{Wat. Scheme}} & \textbf{Trained SM} & \textbf{Trained SM + UP} & \textbf{Trained SM + TP} &\textbf{Trained SM + WN} \\
        \midrule
        \multirow{6}{*}{\makecell{ARC \\ Challenge \\ (ACC)}} & \multirow{6}{*}{0.3166} & \multirow{3}{*}{KGW} & $n=1$ & 0.3404 & 0.3430 \textcolor[rgb]{0.0,0.5,0.0}{(+0.8\%)} & 0.3259 \textcolor[rgb]{0.5,0.0,0.0}{(-4.3\%)}& \textbf{\underline{0.3660}} \textcolor[rgb]{0.0,0.5,0.0}{(+7.5\%)}\\
        & & & $n=2$ & \textbf{\underline{0.3515}} & 0.3200 \textcolor[rgb]{0.5,0.0,0.0}{(-9.0\%)}& 0.2841 \textcolor[rgb]{0.5,0.0,0.0}{(-19.2\%)}& 0.3498 \textcolor[rgb]{0.5,0.0,0.0}{(-0.5\%)}\\
        & & & $n=3$ & \textbf{\underline{0.3712}} & 0.3072 \textcolor[rgb]{0.5,0.0,0.0}{(-17.2\%)}& 0.2935  \textcolor[rgb]{0.5,0.0,0.0}{(-20.9\%)}& 0.3626 \textcolor[rgb]{0.5,0.0,0.0}{(-2.3\%)}\\
        \cmidrule{3-8}
        & & \multirow{3}{*}{\makecell{SynthID\\ -Text}} & $n=1$ & 0.3541 & 0.3259 \textcolor[rgb]{0.5,0.0,0.0}{(-8.0\%)} & 0.3242 \textcolor[rgb]{0.5,0.0,0.0}{(-8.4\%)}& \textbf{\underline{0.3626}} \textcolor[rgb]{0.0,0.5,0.0}{(+2.4\%)} \\
        & & & $n=2$ & 0.3481 & 0.3345 \textcolor[rgb]{0.5,0.0,0.0}{(-3.9\%)}& 0.3200 \textcolor[rgb]{0.5,0.0,0.0}{(-8.1\%)}&  \textbf{\underline{0.3575}} \textcolor[rgb]{0.0,0.5,0.0}{(+2.7\%)}\\
        & & & $n=3$ & \textbf{\underline{0.3532}} & 0.3396 \textcolor[rgb]{0.5,0.0,0.0}{(-3.9\%)}& 0.3251 \textcolor[rgb]{0.5,0.0,0.0}{(-8.0\%)}&  0.3498 \textcolor[rgb]{0.5,0.0,0.0}{(-1.0\%)}\\

        \midrule
        \multirow{6}{*}{\makecell{TruthfulQA \\ Multiple Choice \\ (ACC)}} & \multirow{6}{*}{0.3768} & \multirow{3}{*}{KGW} & $n=1$ & 0.3783 & 0.3815 \textcolor[rgb]{0.0,0.5,0.0}{(+0.8\%)}& 0.3796 \textcolor[rgb]{0.0,0.5,0.0}{(+0.3\%)} & \textbf{\underline{0.3820}} \textcolor[rgb]{0.0,0.5,0.0}{(+1.0\%)}\\
        & & & $n=2$ & 0.4022 & 0.3719 \textcolor[rgb]{0.5,0.0,0.0}{(-7.5\%)} &  0.3887 \textcolor[rgb]{0.5,0.0,0.0}{(-3.4\%)}& \textbf{\underline{0.4026}} \textcolor[rgb]{0.0,0.5,0.0}{(+0.1\%)}\\
        & & & $n=3$ & 0.4061 & 0.3949 \textcolor[rgb]{0.5,0.0,0.0}{(-2.8\%)}& 0.3862 \textcolor[rgb]{0.5,0.0,0.0}{(-4.9\%)}& \textbf{\underline{0.4411}} \textcolor[rgb]{0.0,0.5,0.0}{(+8.6\%)}\\
        \cmidrule{3-8}
        & & \multirow{3}{*}{\makecell{SynthID\\ -Text}} & $n=1$ & 0.4023 & 0.3869 \textcolor[rgb]{0.5,0.0,0.0}{(-3.8\%)} & 0.3497 \textcolor[rgb]{0.5,0.0,0.0}{(-13.1\%)} & \textbf{\underline{0.4027}} \textcolor[rgb]{0.0,0.5,0.0}{(+0.1\%)} \\
        & & & $n=2$ & 0.3912 & 0.3893 \textcolor[rgb]{0.5,0.0,0.0}{(-0.5\%)}& 0.3921 \textcolor[rgb]{0.0,0.5,0.0}{(+0.2\%)} & \textbf{\underline{0.4177}} \textcolor[rgb]{0.0,0.5,0.0}{(+6.8\%)}\\
        & & & $n=3$ & 0.4140 & 0.3883 \textcolor[rgb]{0.5,0.0,0.0}{(-6.2\%)}&  0.3941 \textcolor[rgb]{0.5,0.0,0.0}{(-4.8\%)}& \textbf{\underline{0.4224}} \textcolor[rgb]{0.0,0.5,0.0}{(+2.0\%)}\\
        
        \midrule
        \multirow{6}{*}{\makecell{MTBench \\ (Full Score: 10)}} & \multirow{6}{*}{2.78} & \multirow{3}{*}{KGW} & $n=1$ & 2.88 & 1.87 \textcolor[rgb]{0.5,0.0,0.0}{(-35.1\%)}& 1.27 \textcolor[rgb]{0.5,0.0,0.0}{(-55.9\%)} & \textbf{\underline{3.20}} \textcolor[rgb]{0.0,0.5,0.0}{(+11.1\%)}\\
        & & & $n=2$ & \textbf{\underline{2.90}} & 1.31 \textcolor[rgb]{0.5,0.0,0.0}{(-54.8\%)}& 1.19 \textcolor[rgb]{0.5,0.0,0.0}{(-59.0\%)}& 2.85 \textcolor[rgb]{0.5,0.0,0.0}{(-1.7\%)}\\
        & & & $n=3$ & \textbf{\underline{2.88}}  & 1.30 \textcolor[rgb]{0.5,0.0,0.0}{(-54.9\%)}& 1.16 \textcolor[rgb]{0.5,0.0,0.0}{(-59.7\%)}& \textbf{\underline{2.88}} \textcolor[rgb]{0.0,0.5,0.0}{(+0.0\%)}\\
        \cmidrule{3-8}
        & & \multirow{3}{*}{\makecell{SynthID\\ -Text}} & $n=1$ & \textbf{\underline{3.21}} & 1.51 \textcolor[rgb]{0.5,0.0,0.0}{(-53.0\%)}& 1.26 \textcolor[rgb]{0.5,0.0,0.0}{(-60.7\%)}& 3.19 \textcolor[rgb]{0.5,0.0,0.0}{(-0.6\%)} \\
        & & & $n=2$ & \textbf{\underline{3.11}} & 1.29 \textcolor[rgb]{0.5,0.0,0.0}{(-58.5\%)} & 1.31 \textcolor[rgb]{0.5,0.0,0.0}{(-57.9\%)} & 2.83 \textcolor[rgb]{0.5,0.0,0.0}{(-9.0\%)}\\
        & & & $n=3$ & \textbf{\underline{3.09}} & 1.28 \textcolor[rgb]{0.5,0.0,0.0}{(-58.6\%)}& 1.27 \textcolor[rgb]{0.5,0.0,0.0}{(-58.9\%)}& 2.98 \textcolor[rgb]{0.5,0.0,0.0}{(-3.6\%)}\\
        \bottomrule
    \end{tabular}
    }
    \label{tab:knowledge_llama3_2}
\end{table*}
Table \ref{tab:knowledge_llama3_2} reveals that Llama-3.2-1b, when trained on 200,000 watermarked samples from the teacher model, achieves substantial performance gains across all benchmarks. As for knowledge preservation performance of the three proposed watermark removal methods, the results of Llama-3.2-1b echo the main experimental results: while both UP and TP lead to widespread performance degradation under most configurations, WN stands out for its remarkable ability to preserve knowledge. Specifically, WN yields performance improvements in roughly half of the settings while showing slight decreases in the remaining scenarios, and these changes remain small throughout.

\section{Experiment Results for More Watermarking Schemes}
\label{sec:more_schemes}
\begin{table}[t]
    \caption{Median p-values for watermark detection using \textbf{WN} (Watermark Neutralization), compared against direct training (No Attack) and unwatermarked conditions (Unw.). \raisebox{0.5ex}{\colorbox{blue!30}{\quad}} indicates high watermark confidence, \raisebox{0.5ex}{\colorbox{blue!10}{\quad}} indicates low watermark confidence, and unshaded cells indicate insufficient evidence for watermark presence. The watermarking schemes used in this table are MinHash and SkipHash.}
    \centering
    \resizebox{0.75\textwidth}{!}{
    \begin{tabular}{cccccc}
        \toprule
        \textbf{Watermarking Scheme} & \textbf{Model} & \textbf{Token Num.} & \textbf{Unw.} & \textbf{No Attack} & \textbf{WN} \\
        \midrule
        \multirow{6}{*}{\makecell{MinHash \\ $n=3$}}& \multirow{3}{*}{Llama-7b} & 100k & 7.87e-01 & \cellcolor{blue!30}2.32e-16 & 8.41e-01 \\
        & & 300k & 9.23e-01 & \cellcolor{blue!30}4.52e-47 & 9.28e-01 \\
        & & 1 million & 9.97e-01 & \cellcolor{blue!30}2.79e-141 & 9.98e-01 \\
        \cmidrule{2-6}
        & \multirow{3}{*}{Llama-3.2-1b} & 100k & 7.87e-01 & \cellcolor{blue!30}6.28e-44 & 9.89e-01 \\
        & & 300k & 9.23e-01 & \cellcolor{blue!30}1.14e-134 & 9.96e-01 \\
        & & 1 million & 9.97e-01 & \cellcolor{blue!30}4.56e-385 & 9.98e-01 \\
        \midrule
        \multirow{6}{*}{\makecell{SkipHash \\ $n=3$}}& \multirow{3}{*}{Llama-7b} & 100k & 9.98e-01 & \cellcolor{blue!10}2.80e-05 & 8.91e-01 \\
        & & 300k & 9.98e-01 & \cellcolor{blue!30}1.85e-13 & 9.84e-01 \\
        & & 1 million & 9.98e-01 & \cellcolor{blue!30}4.13e-39 & 9.92e-01 \\
        \cmidrule{2-6}
        & \multirow{3}{*}{Llama-3.2-1b} & 100k & 9.98e-01 & \cellcolor{blue!30}1.07e-07 & 9.91e-01 \\
        & & 300k & 9.98e-01 & \cellcolor{blue!30}4.01e-18 & 9.97e-01 \\
        & & 1 million & 9.98e-01 & \cellcolor{blue!30}3.45e-47 & 9.98e-01 \\
        \bottomrule
    \end{tabular}
    }
    \label{tab:removal_more_scheme}
\end{table}
\begin{table*}[t]
    \caption{Comparison of student model performance across benchmarks under no attack scenario (Trained SM) and using WN for watermark removal. Values in () indicate percentage changes relative to Trained SM, with the highest performance in each setting bolded and underlined.}
    \centering
    \resizebox{0.85\textwidth}{!}{
    \begin{tabular}{ccccccc}
        \toprule
        \textbf{Benchmark} & \textbf{Student Model} &
        \textbf{Ori. SM} & \multicolumn{2}{c}{\textbf{Wat. Scheme}} & \textbf{Trained SM} &\textbf{Trained SM + WN} \\
        \midrule
        \multirow{4}{*}{\makecell{ARC \\ Challenge \\ (ACC)}} & \multirow{2}{*}{Llama-7b} & \multirow{2}{*}{0.4181} & MinHash & $n=3$ & \textbf{\underline{0.4505}}  & 0.4471 \textcolor[rgb]{0.5,0.0,0.0}{(-0.8\%)}\\
        \cmidrule{4-7}
        & & & SkipHash & $n=3$ & \textbf{\underline{0.4625}} & \textbf{\underline{0.4625}} \textcolor[rgb]{0.0,0.5,0.0}{(+0.0\%)}\\
        \cmidrule{2-7}
        & \multirow{2}{*}{Llama-3.2-1b} & \multirow{2}{*}{0.3166} & MinHash & $n=3$ & 0.3532 & \textbf{\underline{0.3567}} \textcolor[rgb]{0.0,0.5,0.0}{(+1.0\%)}\\
        \cmidrule{4-7}
        & & & SkipHash & $n=3$ & 0.3609 & \textbf{\underline{0.3618}} \textcolor[rgb]{0.0,0.5,0.0}{(+0.2\%)}\\
        \midrule
        \multirow{4}{*}{\makecell{TruthfulQA \\ Multiple Choice \\ (ACC)}} & \multirow{2}{*}{Llama-7b} & \multirow{2}{*}{0.3407} & MinHash & $n=3$ & 0.4836 & \textbf{\underline{0.4872}} \textcolor[rgb]{0.0,0.5,0.0}{(+0.7\%)}\\
        \cmidrule{4-7}
        & & & SkipHash & $n=3$ & 0.4562 & \textbf{\underline{0.4624}} \textcolor[rgb]{0.0,0.5,0.0}{(+1.4\%)}\\
        \cmidrule{2-7}
        & \multirow{2}{*}{Llama-3.2-1b} & \multirow{2}{*}{0.3768} & MinHash & $n=3$ & 0.4153 & \textbf{\underline{0.4187}} \textcolor[rgb]{0.0,0.5,0.0}{(+0.8\%)}\\
        \cmidrule{4-7}
        & & & SkipHash & $n=3$ & 0.4037 &  \textbf{\underline{0.4069}} \textcolor[rgb]{0.0,0.5,0.0}{(+0.8\%)}\\
        \midrule
        \multirow{4}{*}{\makecell{MTBench \\ (Full Score: 10)}} & \multirow{2}{*}{Llama-7b} & \multirow{2}{*}{2.64} & MinHash & $n=3$ & \textbf{\underline{3.75}} & 3.73 \textcolor[rgb]{0.5,0.0,0.0}{(-0.5\%)}\\
        \cmidrule{4-7}
        & & & SkipHash & $n=3$ & \textbf{\underline{4.11}} & 4.04 \textcolor[rgb]{0.5,0.0,0.0}{(-1.7\%)}\\
        \cmidrule{2-7}
        & \multirow{2}{*}{Llama-3.2-1b} & \multirow{2}{*}{2.78} & MinHash & $n=3$ & \textbf{\underline{3.03}}  & 2.91 \textcolor[rgb]{0.5,0.0,0.0}{(-4.0\%)}\\
        \cmidrule{4-7}
        & & & SkipHash & $n=3$ & \textbf{\underline{3.46}} & 3.37 \textcolor[rgb]{0.5,0.0,0.0}{(-2.6\%)}\\
        \bottomrule
    \end{tabular}
    }
    \label{tab:knowledge_more_scheme}
\end{table*}
The main experiments focused on two representative watermarking schemes, KGW \cite{DBLP:conf/icml/KirchenbauerGWK23} and SynthID-Text \cite{Dathathri2024}, for testing. This section presents supplementary experiments with additional n-gram based watermarking schemes to demonstrate the strong generalization capability of the proposed method. Moreover, we will discuss several other watermarking paradigms that represent alternative approaches in this field. 

\subsection{Experiment Results for More N-gram based Watermarking Schemes}
\label{sec:more_n_gram}
\vspace{3pt}

\noindent\textbf{MinHash} \quad This method is a variant of KGW, proposed by \citet{kirchenbauer2023reliability}. In the default implementation of KGW, the hash function $H$ is a multiplicative modular function, expressed as:
\begin{equation}
h_t = H(x_{t-n+1:t-1}) = \prod_{i=t-n+1}^{t-1} x_i \bmod |\mathcal{V}|.
\end{equation}
This approach causes the hash result to change whenever any token within the window is modified, leading to reduced robustness as the window size $n$ increases. To address this limitation, several improved versions have been proposed, including MinHash, which uses the minimum token id within the window as the hash result:
\begin{equation}
h_t = H(x_{t-n+1:t-1}) = \min_{i \in [t-n+1,t-1]} x_i
\end{equation}

\vspace{3pt}

\noindent\textbf{SkipHash} \quad SkipHash \cite{kirchenbauer2023reliability} is also a variant of KGW designed to improve robustness, but it uses a hash function that takes the leftmost token id within the window, expressed as:
\begin{equation}
h_t = H(x_{t-n+1:t-1}) = x_{t-n+1}
\end{equation}

\vspace{3pt}

When $n \leq 2$, MinHash and SkipHash are equivalent to KGW, so we only evaluate scenarios where $n=3$. Given WN's superior overall performance among the three proposed methods in terms of watermark removal effectiveness and knowledge preservation, the subsequent experiments exclusively focus on this approach. The experimental results are shown in Table \ref{tab:removal_more_scheme} and Table \ref{tab:knowledge_more_scheme}. The observed trends in the experimental results align consistently with the main experiment, which uses KGW and SynthID-Text. The WN approach demonstrates complete watermark removal efficacy while showing no significant impact on the knowledge acquired by the student model.

\subsection{Discussion About Other Watermarking Paradigms}
\label{sec:more_para}
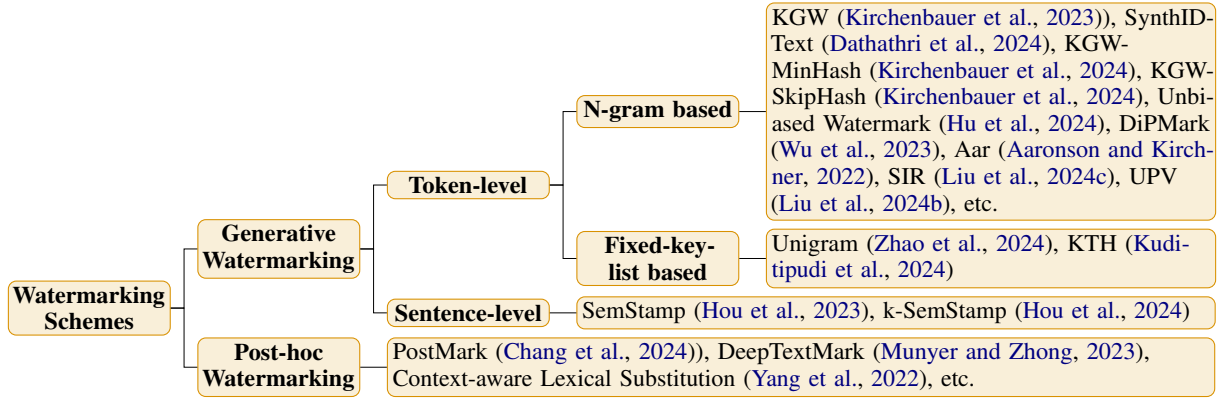
\begin{figure}[h!]
\centering
\tikzset{
        my node/.style={
            draw,
            align=center,
            thin,
            text width=1.2cm, 
            rounded corners=3,
        },
        my leaf/.style={
            draw,
            align=left,
            thin,
            text width=8.5cm, 
            rounded corners=3,
        }
}
\forestset{
  every leaf node/.style={
    if n children=0{#1}{}
  },
  every tree node/.style={
    if n children=0{minimum width=1em}{#1}
  },
}
\begin{forest}
    nonleaf/.style={font=\small},
     for tree={%
        every leaf node={my leaf, font=\small},
        every tree node={my node, font=\small, l sep-=4.5pt, l-=1.pt},
        anchor=west,
        inner sep=2pt,
        l sep=10pt, 
        s sep=3pt, 
        fit=tight,
        grow'=east,
        edge={ultra thin},
        parent anchor=east,
        child anchor=west,
        if n children=0{}{nonleaf}, 
        edge path={
            \noexpand\path [draw, \forestoption{edge}] (!u.parent anchor) -- +(5pt,0) |- (.child anchor)\forestoption{edge label};
        },
        if={isodd(n_children())}{
            for children={
                if={equal(n,(n_children("!u")+1)/2)}{calign with current}{}
            }
        }{}
    }
    [\textbf{Watermarking Schemes}, draw=harvestgold, fill=harvestgold!15, text width=2cm, text=black
        [\textbf{Generative Watermarking}, color=harvestgold, fill=harvestgold!15, text width=2cm, text=black
                [\textbf{Token-level}, color=harvestgold, fill=harvestgold!15, text width=2cm, text=black
                [\textbf{N-gram based},  color=harvestgold, fill=harvestgold!15, text width=2cm, text=black
                    [{KGW \cite{DBLP:conf/icml/KirchenbauerGWK23}), SynthID-Text \cite{Dathathri2024}, KGW-MinHash \cite{kirchenbauer2023reliability}, KGW-SkipHash \cite{kirchenbauer2023reliability}, Unbiased Watermark \cite{hu2023unbiased}, DiPMark \cite{wu2023dipmark}, Aar \cite{aronsonpowerpoint}, SIR \cite{liu2024a}, UPV \cite{liu2024an}, etc.}, color=harvestgold, fill=harvestgold!15, text width=5.8cm, text=black]
                ]
                [\textbf{Fixed-key-list based},  color=harvestgold, fill=harvestgold!15, text width=2cm, text=black
                [{Unigram \cite{zhao2023provable}, KTH \cite{kuditipudi2023robust}}, color=harvestgold, fill=harvestgold!15, text width=5.8cm, text=black]
                ]
                ]
                [
                \textbf{Sentence-level}, color=harvestgold, fill=harvestgold!15, text width=2cm, text=black
                [{SemStamp \cite{hou2023semstamp}, k-SemStamp \cite{hou-etal-2024-k}}, color=harvestgold, fill=harvestgold!15, text width=8.3cm, text=black
                ]
                ]
        ]
        [\textbf{Post-hoc Watermarking}, color=harvestgold, fill=harvestgold!15, text width=2cm, text=black
            [{PostMark \cite{chang2024postmark}), DeepTextMark \cite{munyer2023deeptextmark}, Context-aware Lexical Substitution \cite{yang2022tracing}, etc.}, color=harvestgold, fill=harvestgold!15, text width=10.8cm, text=black
        ]
    ]
]
]
\end{forest}
\caption{Taxonomy of existing watermarking schemes.}
\label{fig:taxonomy}
\end{figure}

As illustrated in Figure \ref{fig:taxonomy}, current watermarking schemes can be categorized into two main approaches: generative watermarking, where watermarks are embedded during the text generation process, and post-hoc watermarking, where watermarks are added to existing texts. Within generative watermarking, there are further subdivisions into token-level methods and sentence-level approaches based on reject sampling. In real-time services, tokens can be outputted as they are sampled while inference continues, thereby enhancing user experience. Post-hoc watermarking and sentence-level watermarking introduce significant latency, making them less suitable for real-time LLM services compared to token-level watermarking. 

Among token-level methods, the predominant paradigm is the n-gram based approach. Additionally, there are few methods that employ a fixed-key-list based approach, which utilizes global fixed watermark keys independent of the prefix. These methods include Unigram \cite{zhao2023provable} and KTH \cite{kuditipudi2023robust}, for which we conduct supplementary experiments. 

\vspace{3pt}

\textbf{Unigram} \cite{zhao2023provable} \quad This approach is equivalent to KGW with prefix length of 0, using a globally fixed red-green partition. According to our main experimental results, both TP and WN can completely remove the watermarks inherited by the student model.

\vspace{3pt}

\textbf{KTH} \cite{kuditipudi2023robust} \quad This method employs a globally fixed sequence of watermark keys: $\xi = \xi^{(0)}, \xi^{(1)}, ...,\xi^{(m-1)}$, where each $\xi^{(j)}\in [0,1]^{|\mathcal{V}|}$ follows a uniform distribution. During text generation, first a random shift $s \in [0,m)$ is selected, then the $t$-th generated token is chosen using the following strategy:
\begin{equation}
    x_t = \mathop{\arg\max}\limits_{i} (\xi_i^{(s+t\mod m)})^{1/p_t},
\end{equation}
where $p_t$ is the original probability prediction at position $t$. During watermark detection, it computes the minimum Levenshtein distance $d$ between the text to be detected $x$ and the key sequences $\xi$. In comparison, it randomly generates $n$ sequences with the same shape as $x_i$, and calculates $d'_1$, $d'_2$, ..., $d'_n$ using the same method. The detection p-value is represented by the proportion of values in the $d'$ sequence that are lower than $d$. It is worth noting that the p-value of this detection method is bounded by the number of trials $n$.

\begin{table}[h!]
\caption{Median p-value of watermark detection in trained student models using KTH watermarking scheme ($m=256$, token num. = 256, $n=100$).}
\centering
\resizebox{0.7\textwidth}{!}{
\begin{tabular}{ccc}
\toprule
\textbf{Watermarking Scheme} & \textbf{Student Model} & \textbf{Median p-value (No Attack)} \\
\midrule
\multirow{2}{*}{KTH} & Llama-7b & 3.1e-01 \\
& Llama-3.2-1b & 3.2e-01 \\
\bottomrule
\end{tabular}
}
\label{tab:kth}
\end{table}

Our experiments revealed that the addition of KTH watermark significantly affects the instruction-following capability of the teacher model, resulting in a lower proportion of QA pairs conforming to format rules and generally shorter answers. After training on such data, the watermark is barely detectable in the student model, as shown in Table \ref{tab:kth}.

\subsection{Discussion About Repeated Context Masking}
Another noteworthy watermarking technique to examine is repeated context masking, where watermarks are only applied to subsequent tokens when a context is first encountered, but not when similar contexts appear later in the sequence. This approach has been implemented in several watermarking algorithms, including SynthID-Text \cite{Dathathri2024} (which offers both a distortionary version and a non-distortionary version utilizing repeated context masking, with our main experiments employing the distortionary variant) and Unbiased watermark \cite{hu2023unbiased}. The technique is specifically designed to ensure the watermarking process remains distortion-free.

Using repeated context masking could significantly reduce watermark radioactivity. As demonstrated in Figure \ref{fig:freq_learnability}, watermark radioactivity exhibits a strong correlation with prefix frequency in the training data. When repeated context masking is implemented, the frequency of watermark patterns in the training data diminishes substantially, thereby impeding the student models' ability to acquire these patterns.

We conducted experimental evaluations of both SynthID-Text and Unbiased Watermark employing repeated context masking, while maintaining other parameters consistent with the main experiments. In the implementation of repeated context masking, we established a maximum storage capacity of 1024 distinct contexts. For student models without any watermark removal attack, the median p-value of watermark detection are shown in Table \ref{tab:repeated_context_masking}.
\begin{table*}[h!]
\caption{Median p-value of watermark detection in student models trained on texts watermarked with repeated context masking. }
\centering
\resizebox{0.87\textwidth}{!}{
\begin{tabular}{ccccc}
\toprule
\multicolumn{2}{c}{\textbf{Watermarking Scheme}} & \textbf{Token Num}. & \textbf{Llama-7b} & \textbf{Llama-3.2-1b} \\
\midrule
\multirow{6}{*}{{SynthID-Text (w. Context Mask)}}& \multirow{3}{*}{$n=2$} & 10k & 1.50e-01 & 1.10e-01 \\
& & 20k & 1.00e-01 & 8.34e-02 \\
& & 30k & 3.00e-02 & 2.11e-02 \\
\cmidrule{2-5}
& \multirow{3}{*}{$n=3$} & 100k & 3.50e-01 & 2.80e-01 \\
& & 300k & 2.80e-01 & 2.20e-01 \\
& & 1 million & 2.10e-01 & 1.70e-01 \\
\midrule
\multirow{6}{*}{{UW (w. Context Mask)}}& \multirow{3}{*}{$n=2$} & 10k & 2.10e-01 & 1.80e-01 \\
& & 20k & 1.60e-01 & 1.30e-01 \\
& & 30k & 9.50e-02 & 7.20e-02 \\
\cmidrule{2-5}
& \multirow{3}{*}{$n=3$} & 100k & 4.50e-01 & 3.80e-01 \\
& & 300k & 3.70e-01 & 3.00e-01 \\
& & 1 million & 3.00e-01 & 2.50e-01 \\
\bottomrule
\end{tabular}
}
\label{tab:repeated_context_masking}
\end{table*}

\section{Impact of Inference-Time Watermark Neutralization (WN) on Knowledge Preservation under Non-watermarked Setting}
Previous experiments have demonstrated that when teacher model is watermarked, inference-time watermark neutralization (WN) effectively enables the student model to bypass the watermark while acquiring knowledge from the teacher model's outputs that is comparable to what would be learned without any attack. In this section, we conducted additional experiments to examine whether applying WN would affect the knowledge acquired by the student model in cases where the teacher model itself is not watermarked (noting that the student model has no access to the detector and thus cannot determine the presence of watermarks).

Consistent with the settings in our main experiments, we generated QA pairs using GLM-4-9b-chat. After filtering and deduplication, we obtained 200,000 non-watermarked samples. We trained both Llama-7b and Llama-3.2-1b models using this dataset, and then applied WN for watermark removal and evaluated the performance changes across various benchmarks, with results presented in Table \ref{tab:knowledge_non_wat}. It can be concluded that WN \textit{does not} exert a substantial negative impact on knowledge preservation under non-watermarked setting.

\begin{table*}[h!]
    \caption{Impact of WN on knowledge preservation under non-watermarked setting.}
    \centering
    \resizebox{0.78\textwidth}{!}{
    \begin{tabular}{ccccc}
        \toprule
        \textbf{Benchmark} & \textbf{Student Model} &
        \textbf{Ori. SM} & \textbf{Trained SM} & \textbf{Trained SM + WN} \\
        \midrule
        \multirow{2}{*}{\makecell{ARC Challenge \\ (ACC)}} & Llama-7b & 0.4181 & 0.4454 & 0.4488 \textcolor[rgb]{0.0,0.5,0.0}{(+0.7\%)}\\
        & Llama-3.2-1b & 0.3166 & 0.3524 & 0.3507 \textcolor[rgb]{0.5,0.0,0.0}{(-0.5\%)}\\
        \midrule
        \multirow{2}{*}{\makecell{TruthfulQA \\ Multiple Choice (ACC)}} & Llama-7b & 0.3407 & 0.4254 & 0.4507 \textcolor[rgb]{0.0,0.5,0.0}{(+5.9\%)}\\
        & Llama-3.2-1b & 0.3768 & 0.4052 & 0.3951 \textcolor[rgb]{0.5,0.0,0.0}{(-2.5\%)}\\
        \midrule
        \multirow{2}{*}{\makecell{MTBench \\ (Full Score: 10)}} & Llama-7b & 2.64 & 3.99 & 4.12 \textcolor[rgb]{0.0,0.5,0.0}{(+3.2\%)}\\
        & Llama-3.2-1b & 2.78 & 3.15 & 3.08 \textcolor[rgb]{0.5,0.0,0.0}{(-2.2\%)}\\
        \bottomrule
    \end{tabular}
    }
    \label{tab:knowledge_non_wat}
\end{table*}

\section{Adaptive Control for Inverse Watermark Strength}
\label{sec:adaptive}
Throughout all previous experiments, we consistently used an inverse watermark strength of $\delta'=2.5$ for WN, which achieved complete watermark removal in all cases. As detailed in Appendix \ref{sec:schemes}, the chosen watermark strength of the teacher model represents a notably high intensity that remains practical for deployment in LLM services, suggesting that $\delta'=2.5$ is sufficient for the vast majority of scenarios.

However, to account for potential extreme cases, we also explored strategies for adaptive control of inverse watermark strength. Our approach is to estimate the required inverse watermark strength $\delta'$ by detecting the watermark intensity inherited by the student model. Since the student model holder does not have access to the watermark detector, we employed Water-Probe \cite{liu2024can} to measure watermark intensity. Water-Probe is a recently proposed identification algorithm that tests for watermarks by comparing the model's responses to specially crafted prompts, where higher similarity in responses to crafted prompts pairs indicates a higher likelihood (or strength) of watermarking.

We conducted experiments using KGW with $n=1$. Table \ref{tab:waterprobe} shows the cosine similarity scores detected by Water-Probe-v2\footnote{There are two versions of Water-Probe, with version 2 demonstrating more stable performance in our experiments.} for Llama-7b student models trained with different watermark strengths $\delta$, compared with a student model trained on unwatermarked data. Based on this reference table, we can estimate the watermark strength $\delta$ used in the teacher model by examining the WaterProbe-v2 cosine similarity score of the trained student model. This estimation enables us to adaptively select an appropriate inverse watermark strength $\delta'$ for removal. 

Here is a practical example: Suppose the teacher model is watermarked using KGW with $n=1, \delta=5.0$. The trained student model's detected cosine similarity is 0.1694, which is slightly higher than the reference value of 0.1366 for $\delta=3.0$. Given our prior knowledge that $\delta'=2.5$ can completely remove watermarks with $\delta=3.0$, we should proportionally increase the inverse watermark strength. Therefore, we set $\delta'=3.0$ for this case. The removal results are shown in Table \ref{tab:kgw_delta5}.

\begin{table}[t]
\caption{Water-Probe-v2 cosine similarity scores for student models under different watermark strength settings.}
\centering
\begin{tabular}{lcccccc}
\toprule
\multirow{2}{*}{\textbf{Settings}} & \multirow{2}{*}{Unw.} & \multicolumn{5}{c}{KGW with Different $\delta$} \\
\cmidrule(lr){3-7}
 & & $\delta=1.0$ & $\delta=2.0$ & $\delta=3.0$ & $\delta=4.0$ & $\delta=5.0$ \\
\midrule
\textbf{Cosine Similarity} & 0.0065 & 0.0853 & 0.1032 & 0.1366 & 0.1544 & 0.1694 \\
\bottomrule
\end{tabular}
\label{tab:waterprobe}
\end{table}
\begin{table}[t]
\caption{Median p-values for watermark detection using WN under different inverse watermark strengths. The watermark used in teacher model is KGW, with $\delta=5.0$, the student model is Llama-7b.}
\centering
\resizebox{0.55\textwidth}{!}{
\begin{tabular}{cccc}
\toprule
\textbf{Token Number} & \textbf{No Attack} & \textbf{WN $\delta'=2.5$} & \textbf{WN $\delta'=3.0$} \\
\midrule
10k & \cellcolor{blue!30}7.75e-59 & \cellcolor{blue!10}2.81e-03 & 9.17e-02 \\
20k & \cellcolor{blue!30}7.48e-115 & \cellcolor{blue!10}6.91e-05 & 4.29e-02 \\
30k & \cellcolor{blue!30}1.14e-170 & \cellcolor{blue!30}6.31e-07 & 1.13e-02 \\
\bottomrule
\end{tabular}
}
\label{tab:kgw_delta5}
\end{table}

We acknowledge that the current estimation method is relatively rough. However, it's important to emphasize that in practical LLM services, it would be unrealistic to use such strong watermarks as $\delta=5.0$, as this would significantly degrade the output quality. In most cases, selecting an inverse watermark strength of $\delta'=2.5$ is already sufficient.
\section{Details of Training Data Collection}
\label{sec:training-data}
\subsection{Prompt Used for Training Data Collection} 
\label{sec:prompt}
Following the work by \citet{sander2024watermarking}, we prompted the teacher model to generate question-answering samples consisting of instruction, input and answer, as shown in the prompt template in Figure \ref{fig:prompt}.
\begin{figure*}[h!]
\begin{tcolorbox}[colback=gray!10, colframe=black, rounded corners]
You are asked to come up with a set of 20 diverse task instructions and their answers. These instructions will be given to large
language model and we will evaluate it for completing the instructions. Here are the requirements:

\vspace{5pt}

1. Try not to repeat the verb for each instruction to maximize diversity.

2. The language used for the instruction also should be diverse. For example, you should combine questions
with imperative instructions.

3. The type of instructions should be diverse. The list should include diverse types of tasks like open-ended
generation, classification, math reasoning, code generation, world common knowledge, reading comprehension, editing, etc.

2. A language model should be able to complete the instruction. For example, do not ask the assistant to create any visual or audio output. For another example, do not ask the assistant to wake you up at 5pm or set a reminder because it cannot perform any action.

3. The instructions should be in English.

4. The instructions should be 1 to 2 sentences long. Either an imperative sentence or a question is permitted.

5. You should generate an appropriate input to the instruction. It should contain a specific example for the instruction when needed.

6. Most of the instructions do not require input. In this case, you must simply put "None" in the input field.

7. The input should not be the answer to the instruction. For example, if the instruction is "Write a headline for ...", the input should not be the said headline. Instead, the input should be "None".

8. The output should be an appropriate response to the instruction and the input. Make sure the output is less than 100 words.

9. Do not repeat instructions.

10. Use [[, ]] to flag the start and end of each example.

\vspace{5pt}

Examples of generating 3 task instruction and its answer:

\vspace{5pt}

[[Instuction: Give a detailed description of a method for brainstorming an essay.

Input:

Answer: Mind Mapping is one of the methods for brainstorming an essay. Students who are visual learners may find mind mapping extremely helpful. In order to Mind Mapping: Take a paper and draw a circle in the center Write your topic inside it. Just like we sketch rays protruding from a sun, draw lines coming out of the circle. Write an argument or a statement related to the topic at the
end of each line. Draw another set of lines projecting from these statements and state examples that support them. Continue this exercise until you run out of all possibilities imaginable for this topic.]]

\vspace{5pt}

[[Instuction: Write a creative or imaginative response to the given prompt.

Input: You are a time traveler who has just arrived in ancient Egypt. What do you see?

Answer: Wow! As I step out of my time machine, I’m surrounded by towering pyramids and the warm sun beats down on my face. I see Pharaohs riding in grand chariots, and workers building majestic temples. It’s like something out of a dream!]]

\vspace{5pt}

[[Instuction: Solve a math problem.

Input: A snail is at the bottom of a 20-foot well. Each day, it climbs up 3 feet, but at night, it slips back 2 feet. How many days will it take for the snail to reach the top of the well?

Answer: To solve this problem, we first calculate the net distance the snail covers each day: 3 feet up minus 2 feet down, 
resulting in 1 foot of progress daily. Since the well is 20 feet deep, we can set up the equation: number of days $\times$ 1 foot = 20 feet.  Solving this, we find it takes 20 days for the snail to reach the top. We can verify this by noting that after 19 days, the snail will 
have climbed 19 feet. On the 20th day, it will climb the final 3 feet, reaching the top before any night-time slipping occurs. 
Therefore, the snail will reach the top of the well in 20 days.]]
\end{tcolorbox}
\caption{Prompt used for acquiring training data.}
\label{fig:prompt}
\end{figure*}

\subsection{Example Samples of Training Data}
\label{sec:examples}
Example samples of training data produced using different watermarking schemes and $n$ are shown in Figure \ref{fig:training_data}.
\begin{figure*}[h!]
\begin{tcolorbox}[colback=gray!10, colframe=black, rounded corners]
\textbf{KGW $n=1$} \\[0.5em]
Instruction: How does photosynthesis work? \\
Input: None \\
Answer: Photosynthesis is the process by which green plants, algae, and some bacteria use light to synthesize foods with the help of chlorophyll. \\

\textbf{KGW $n=2$} \\[0.5em]
Instruction: Explain how the Internet has transformed the way people learn.\\
Input: None\\
Answer: The Internet has transformed learning by providing access to vast information, allowing flexible learning times, and facilitating interactive communication between learners and instructors.\\

\textbf{KGW $n=3$} \\[0.5em]
Instruction: Provide the definition of a simile.\\
Input: None.\\
Answer: A simile is a figure of speech in which two things are compared using like or as, often used to enhance the vividness of a comparison.\\

\textbf{SynthID-Text $n=1$} \\[0.5em]
Instruction: How do you brew a cup of green tea? \\
Input: None \\
Answer: 1. Boil water: Start by boiling water to the ideal temperature of 160 degrees Fahrenheit (70 degrees Celsius). 2. Steep the tea: Place a green tea bag or loose-leaf green tea in a tea pot or cup. 3. Add hot water: Pour the hot water directly over the tea. 4. Steep for the appropriate time: Green tea typically needs to steep for about 2-3 minutes. 5. Remove the tea: Take the tea bag or strain the loose leaves from the water. 6. Sweeten if desired: Add honey, sugar, or a sweetener of your choice to taste. 7. Serve: Pour the brewed tea into a cup and enjoy.\\

\textbf{SynthID-Text $n=2$} \\[0.5em]
Instruction: Provide a list of three common kitchen tools and their uses. \\
Input: None \\
Answer: 1. Cutting board: A cutting board is used for chopping, slicing, and dicing various ingredients, such as vegetables, fruits, and meats. 2. Measuring cups and spoons: These tools are used for accurately measuring ingredients, ensuring consistent and precise measurements in recipes. 3. Kitchen knife: A kitchen knife is versatile and used for slicing, dicing, chopping, and mincing a wide range of ingredients.\\

\textbf{SynthID-Text $n=3$} \\[0.5em]
Instruction: Analyze the given quote and identify the underlying theme.\\
Input: The only way to do great work is to love what you do.\\
Answer: 1. Passion and joy in work 2. Importance of personal fulfillment in work 3. The necessity of loving one's job\\

\end{tcolorbox}
\caption{Selected examples in training dataset.}
\label{fig:training_data}
\vspace{-15pt}
\end{figure*}

\end{document}